%% file: main.tex
\documentclass{article} 
\usepackage{iclr2027_conference,times}

\input{math_commands.tex}

\usepackage{amsmath,amssymb}
\usepackage{hyperref}
\usepackage{graphicx}
\usepackage{url}

\usepackage[capitalize,noabbrev]{cleveref}

\usepackage{multirow}
\usepackage{xcolor}
\usepackage{colortbl}
\usepackage{booktabs}
\usepackage{float}
\usepackage{wrapfig}
\usepackage{soul}

\usepackage{siunitx}
\title{\raggedright AReA: Adaptive Receptive-Field Attention for Efficient High-Resolution Diffusion Transformers\vspace{-0.1in}}

\author{Javad Rajabi\thanks{Equal contribution. Correspondence: \texttt{rajabi@cs.toronto.edu}}\hspace{0.45em}, Koorosh Roohi$^{*}$, David B.~Lindell, Babak Taati\\[0.65mm]
University of Toronto \quad Vector Institute\\[1mm]
\small Project page: \url{https://rajabi2001.github.io/area/}
\vspace{-3pt}
}

\iclrfinalcopy 

\begin{document}
\vspace*{-0.45in}
\maketitle
\lhead{Preprint. Under review.}


\input{sections/abstract}


\input{sections/intro}

\input{sections/related_work}

\input{sections/method}

\input{sections/experiments}

\input{sections/conclusion}






\bibliography{iclr2027_conference}
\bibliographystyle{iclr2027_conference}

\newpage

\appendix

\input{sections/appendix}

\end{document}

%% file: math_commands.tex
\usepackage{amsmath,amsfonts,bm}

\def\eqref#1{equation~\ref{#1}}

\def\1{\bm{1}}

\DeclareMathAlphabet{\mathsfit}{\encodingdefault}{\sfdefault}{m}{sl}
\SetMathAlphabet{\mathsfit}{bold}{\encodingdefault}{\sfdefault}{bx}{n}



%% file: sections/abstract.tex
\vspace{-25pt}
\begin{figure}[H]
  \centering
  \includegraphics[width=\textwidth]{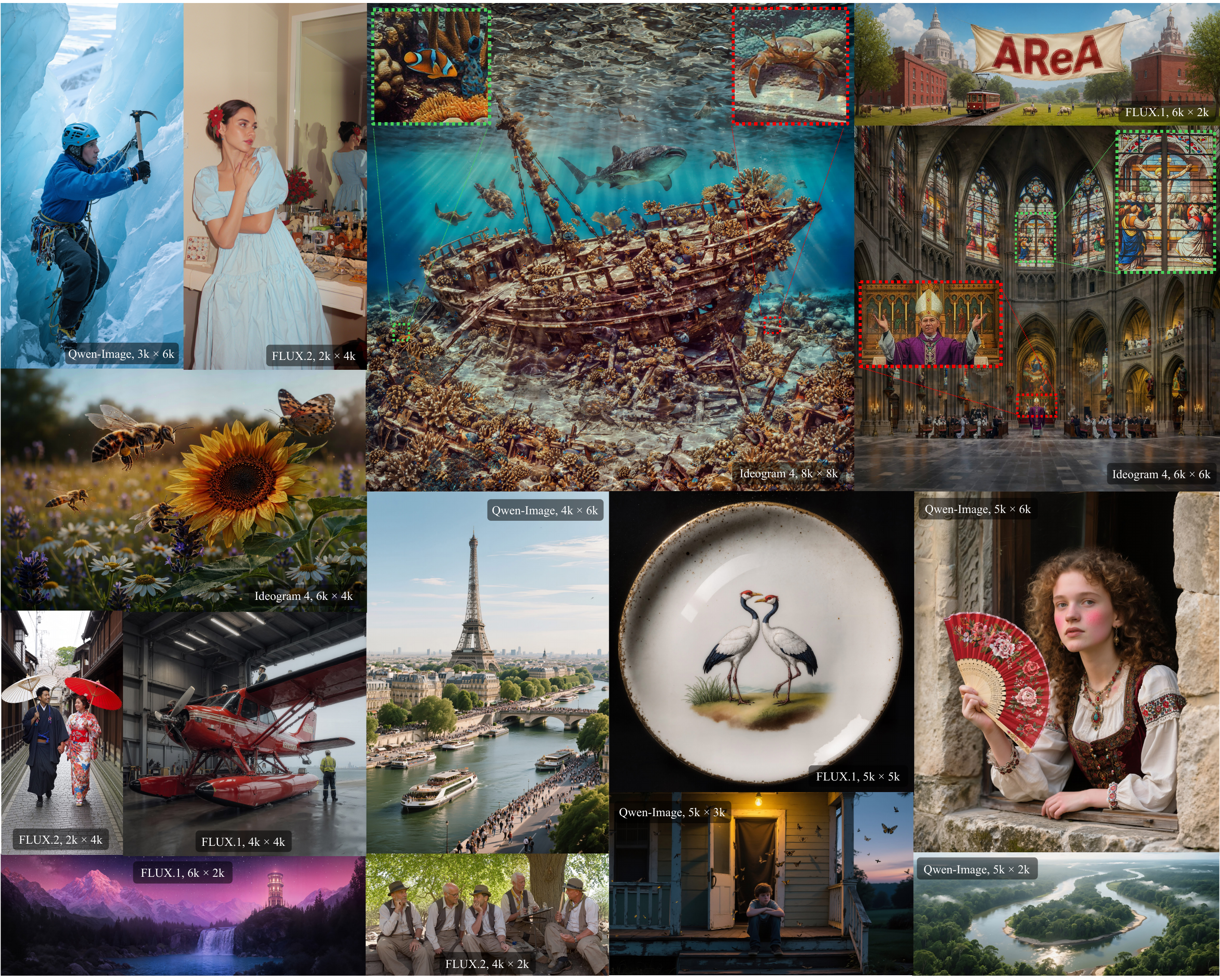}
  \vspace{-20pt}
  \caption{\textbf{High-Resolution Generation.} AReA enables training-free and efficient high resolution generation across FLUX.1, Qwen-Image, FLUX.2, and Ideogram 4. Best viewed zoomed-in.
  \vspace{-10pt}
  }
  \label{fig:teaser}
\end{figure}


\begin{abstract}
\vspace{-7pt}

Generating images beyond the training resolution of Diffusion Transformers (DiTs) remains challenging, as the number of tokens grows rapidly with resolution, full attention becomes expensive, and outputs often suffer from structural distortions and degraded detail. We identify uniform full attention as the common source of both problems. Attending to all tokens at every step causes quadratic cost and exposes the model to more tokens and relative distances beyond the training range, pushing the model out of distribution. We observe that attention's effective receptive field narrows during denoising, from global in early steps that form structure to local in later steps that refine details. We propose \textbf{AReA}, a training-free attention mechanism whose receptive field evolves from coarse to fine, following the denoising progression. In early steps, AReA applies cross-scale coarse attention over downsampled keys and values, capturing global context, while in later steps it applies local attention within progressively shrinking windows. Without additional parameters or training, experiments show that AReA consistently outperforms state-of-the-art training-free baselines across target resolutions and different models, while also being faster and more efficient.

\end{abstract}


%% file: sections/intro.tex
\vspace{5pt}
\begin{figure}[t]
  \centering
  \includegraphics[width=\textwidth]{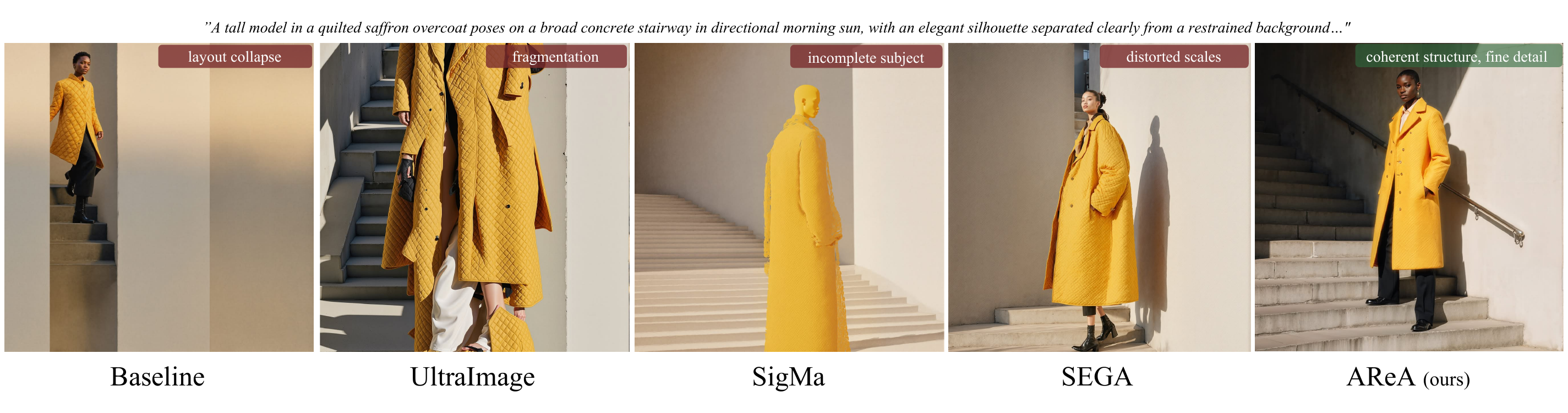}
  \vspace{-22pt}
  \caption{\textbf{Failure Modes at High Resolution}. FLUX.1 at 4096 × 4096. Full attention keeps the model's intended subject but collapses the layout, while RoPE-based methods keep a layout but fragment, lose, or distort the subject. AReA preserves both and renders fine details.
  \vspace{-10pt}
  }
  \label{fig:motives}
\end{figure}

\section{Introduction}
\label{sec:intro}
\vspace{-5pt}

Text-to-image (T2I) generation has witnessed remarkable progress driven by diffusion models~\citep{ddpm, dhariwal2021diffusion}, with Diffusion Transformers (DiTs)~\citep{dit, bao2023all} pushing generation quality even further through highly scalable architectures. Despite these advancements, generating high-resolution images beyond a model's training resolution remains a significant challenge, often leading to noticeable degradation or even complete failure. While training at the target resolution seems to be the obvious remedy, it typically demands massive amounts of high-quality data and substantial computational resources. As a result, a growing body of research has shifted toward training-free approaches, seeking to harness the high-resolution capabilities already embedded in pre-trained diffusion models~\citep{sega, scalediff}.

Existing training-free approaches largely follow two lines. The first adopts a two-scale paradigm, generating a low-resolution image and using it to guide high-resolution synthesis~\citep{hiflow, demofusion, imax, diffusehigh}. While this preserves global structure, the output is limited by the low-resolution guidance rather than by the model’s capacity, often resulting in low-quality images. The second generates at the target resolution in a single end-to-end pass by modifying model components such as positional encodings or the denoising process itself~\citep{ultraimage, dype, sega}. Although this direction seems promising, existing methods are often computationally expensive or tied to architecture-specific components such as Rotary Position Embeddings (RoPE)~\citep{rope}, limiting their generalizability. Moreover, these methods often still produce images with structural distortions (Figure~\ref{fig:motives}). Therefore, efficient generation of high-quality, structurally coherent images across diverse high resolutions remains an open problem.

\begin{figure}[t]
  \centering
  \includegraphics[width=\textwidth]{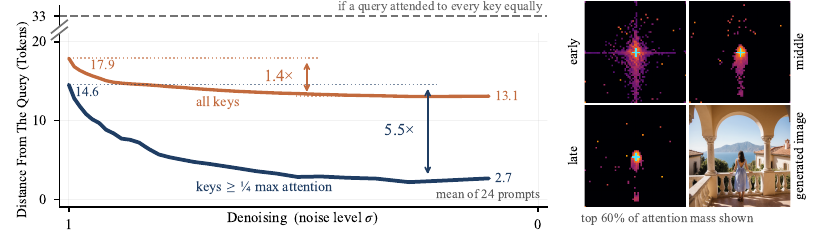}
  \vspace{-17pt}
  \caption{\textbf{Attention Locality as Denoising Progresses.} 
  Vanilla FLUX.1 ($1024 \times 1024$), averaged across heads, blocks, and 24 prompts. \emph{Left}: attention-weighted query–key distance. Over all keys it shrinks only $1.4\times$, but over dominant keys ($\geq 1/4$ max attention) it shrinks $5.5 \times$: late in denoising, the most important keys lie close to the query. \emph{Right}: attention maps of a center query at early, middle, and late steps (top 60\% of mass).
  \vspace{-10pt}
  }
  \label{fig:locality}
\end{figure}

We argue that both the inefficiency and the artifacts share the same root in how tokens interact in attention. At high resolution, attention is still applied uniformly, so every token attends to every other token at every denoising step, regardless of what that step is actually resolving. No consideration is given to how far a token really needs to see; each is simply given a global receptive field, and this unrestricted freedom comes at a price. Token interactions grow quadratically with the number of tokens, making generation increasingly slow at high resolution, and they reach across distances beyond the training range, which pushes the model out of distribution. Uniform attention also ignores the coarse-to-fine nature of denoising~\citep{xiao2026spectral}, in which early steps establish global structure and later steps refine local details. Thus, \textit{the receptive field that attention needs varies across denoising steps} (\Cref{fig:locality}). Uniform global attention at every step is therefore unnecessary, and beyond the training range it is actively harmful. The model already has the capacity; what is needed is to align the attention receptive field with this coarse-to-fine progression.

Motivated by this insight, we introduce \textbf{AReA} (\textbf{A}daptive \textbf{Re}ceptive-Field \textbf{A}ttention), a training-free attention mechanism with a \textit{coarse-to-fine receptive field} that evolves across denoising steps. As shown in \Cref{fig:pipeline}, AReA operates in two phases, applying cross-scale coarse attention in early steps and adaptive windowed fine attention in later steps. In the first phase, AReA attends to pooled keys and values placed on a coarse positional grid matched to the model's native training resolution, so every token still sees the whole image, but through a downsampled view that captures global context at low cost. In the second phase, AReA applies attention within local windows that progressively shrink as fine details emerge, so each token interacts mainly with nearby tokens. Rather than tiling the image uniformly, AReA uses a Discrete Wavelet Transform (DWT) map of the current state to split windows over smooth regions into smaller ones, further reducing the attention cost. In this way, AReA avoids the quadratic cost of full-resolution global attention and suppresses long-range interactions when they are unnecessary, keeping the model close to its training regime (see Appendix~\ref{app:ood}). As a result, AReA requires no RoPE extrapolation and avoids its limitations, while leaving the model unchanged apart from what each query attends to.

Extensive experiments show that AReA consistently reduces generation time, with speedups that grow with resolution, reaching 3.2$\times$ over uniform full-resolution global attention at $6144^2$. Moreover, it improves structural coherence and fine-detail fidelity, outperforming baselines across resolution settings and generating high-quality images of up to 67 million pixels, as shown in Figure~\ref{fig:teaser}. AReA requires no retraining or extra parameters and integrates directly into existing DiT models.

%% file: sections/related_work.tex
\vspace{-7pt}
\section{Related work}
\label{sec:related_work}

\vspace{-5pt}
\paragraph{Training-free High-Resolution Image Synthesis.}
Preserving both global structure and fine-grained detail remains an open challenge in high-resolution generation. To address this, recent work~\citep{multidiffusion, fouriscale, upsample-guidance, jin2023training} proposes training-free strategies for extending diffusion models beyond their training resolution. Early approaches explored cascaded generation~\citep{cascaded}, progressive upsampling~\citep{matryoshka}, domain-specific pipelines~\citep{ultrapixel}, and latent-space super-resolution~\citep{lsrna}. Recent methods have largely adopted multi-stage guidance pipelines, where generation is first performed at the model’s native resolution and the resulting image or trajectory is then used to guide sampling at higher resolutions. DiffuseHigh~\citep{diffusehigh} upsamples the base-resolution image and re-denoises it, using the Discrete Wavelet Transform to preserve its structure. I-Max~\citep{imax} uses projected flows from the base-resolution trajectory to stabilize the formation of coarse structure. Similarly, HiFlow~\citep{hiflow} constructs a virtual reference flow from base-resolution information to guide high-resolution sampling. ScaleDiff~\citep{scalediff} follows a similar paradigm, refining the upsampled base-resolution image with SDEdit~\citep{sdedit} while combining patch-level attention, latent frequency mixing, and structure guidance to balance global structure and fine detail. While effective, these methods inherently rely on base-resolution guidance and introduce additional complexity into the high-resolution denoising process.

\vspace{-5pt}
\paragraph{Direct High-Resolution Inference.}
More recent work aims to unlock the high-resolution capabilities already present in pretrained models, generating images directly at the target resolution by adjusting components such as attention and positional encodings at inference time. For example, ResDiT~\citep{resdit} applies global attention with scaled positional embeddings in parallel with patch-wise attention, then fuses their outputs through Gaussian-weighted blending and spectral fusion. Another line of work approaches the problem through the lens of RoPE length extrapolation, a problem that closely mirrors context extension in large language models (LLMs)~\citep{yarn, longrope}. These methods typically formulate resolution extrapolation as a recalibration of RoPE’s rotary frequencies so that positions beyond the training range remain within the distribution seen during training. DyPE~\citep{dype} introduces time-aware positional adjustments that evolve across diffusion timesteps. UltraImage~\citep{ultraimage} mitigates repetitive artifacts by shifting the dominant frequency to align with the training resolution and employing entropy-guided attention. SigMa~\citep{sigma} uses a scale-adaptive sigmoid schedule to adjust positional modulation according to the target resolution. SEGA~\citep{sega} analyzes the spectral energy of intermediate latents to dynamically adjust attention scaling, amplifying high-energy frequency bands while suppressing low-energy ones to preserve fine details without compromising global structure. While effective, these methods are tied to RoPE and inherit the limits of length extrapolation, restricting their generalizability and often causing artifacts under large resolution shifts. In contrast, AReA requires neither base-resolution guidance nor RoPE extrapolation. It adapts the attention receptive field across denoising steps, shifting from global to local attention as generation moves from coarse structure to fine detail. This keeps the model close to its training regime, remaining stable far beyond the training resolution while reducing attention cost.

%% file: sections/method.tex
\begin{figure}[t]
  \vspace{-5pt}
  \centering
  \includegraphics[width=\textwidth]{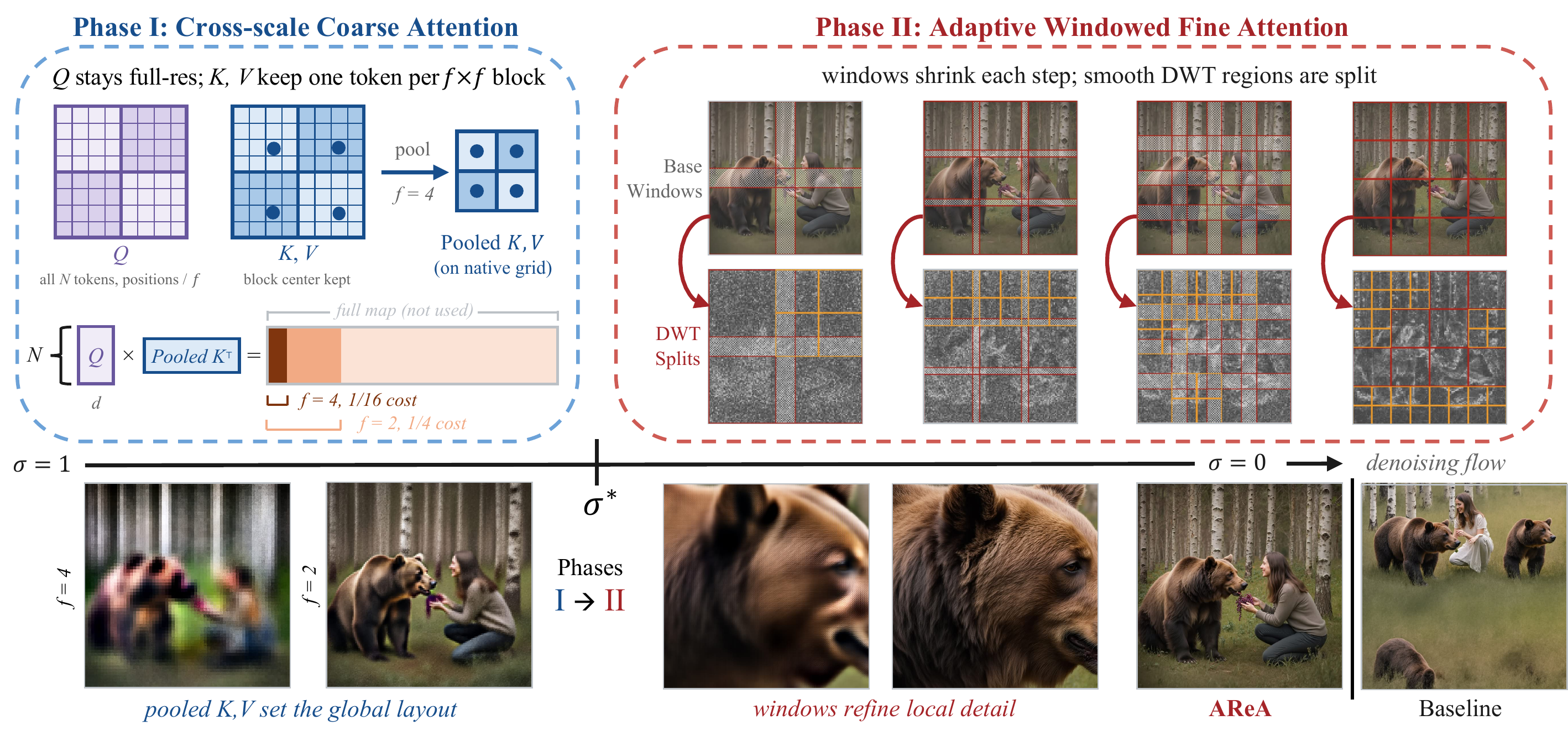}
  \vspace{-20pt}
  \caption{\textbf{Pipeline of AReA.} AReA follows the coarse-to-fine progression of denoising, switching from Phase~I to Phase~II at noise level $\sigma^\star$. In Phase~I, full-resolution queries attend to keys and values pooled to one token per $f \times f$ block, with $f$ ramped over steps, and positioned on a grid matching the training resolution. In Phase~II, attention is restricted to progressively shrinking local windows, and those over smooth regions of the DWT map are further split to reduce cost.
  \vspace{-11pt}
  }
  \label{fig:pipeline}
\end{figure}

\vspace{-6pt}
\section{Method}
\label{sec:method}
\vspace{-6pt}

\textit{Adaptive Receptive-Field Attention} (AReA) is a training-free method
that schedules the attention receptive field by transitioning from a coarse,
global view to fine, local windows, matching the level of detail emerging at
each denoising step. Our approach is grounded in the observation that attention becomes increasingly local as denoising progresses (\Cref{fig:locality}). It is therefore natural to let the attention's receptive field adapt accordingly
rather than remain fixed and uniform across all denoising steps. AReA achieves this in two phases (\Cref{fig:pipeline}). Cross-scale coarse attention first lets full-resolution queries attend to a downsampled view of the image through pooled keys and values, establishing the global layout at low cost, and adaptive windowed fine attention then refines local detail within shrinking windows. As a result, AReA never computes full attention at the target resolution. An important design choice is when to switch between the two phases. Switching too early leaves the composition unresolved, causing inconsistent structure across windows, while switching too late leaves little room for detail refinement and reduces the speedup. Since denoising progress varies across models and samplers, we define the switch by a noise level $\sigma^\star$ rather than a step count, chosen empirically as the point where the composition is established (Appendix~\ref{app:phase-switch}). Phase~I thus covers the first $S_c$ of the $S$ steps, up to $\sigma^\star$, and Phase~II the remaining $S_d = S - S_c$.

\vspace{-4pt}
\subsection{Preliminaries: Attention and Receptive Fields in DiTs}
\vspace{-4pt}

In standard DiTs, an image latent is divided into patches and flattened into a sequence of $N$ tokens, $X \in \mathbb{R}^{N \times D}$. The self-attention layer projects these tokens into Queries ($Q$), Keys ($K$), and Values ($V$). RoPE encodes spatial information by rotating $Q$ and $K$ based on their 2D grid coordinates, enabling attention to capture relative distances. The attention output is then computed as:
\begin{equation}
\text{Attention}(Q, K, V) = \text{softmax}\!\left(\frac{QK^\top}{\sqrt{d}}\right)V,
\end{equation}
where $d$ is the per-head dimension. 
Every query attends to every key, giving each token a global receptive field at $O(N^2)$ cost. While this global reach is needed to establish the layout, it comes at a quadratic cost, much of which may be unnecessary for large $N$. Attention, however, is flexible in what each query attends to. Instead of computing the full $N \times N$ matrix, we can adapt a query’s receptive field by varying its spatial coverage, resolution, or subset of keys and values it attends to. Decoupling queries from keys and values across scales or restricted regions reduces computation while preserving essential interactions. This principle of adaptive receptive fields forms the foundation of AReA’s two-phase schedule.

\vspace{-3pt}
\subsection{Phase I: cross-scale coarse attention}
\label{sec:method:phase1}

The early denoising steps decide the global layout of the image, so each token needs a receptive field that covers the whole grid. Providing it at full target resolution, however, is problematic, as it incurs the quadratic cost we seek to avoid and exposes the model to unseen token counts and relative-position ranges. AReA addresses this via \textit{cross-scale coarse attention}, where queries remain at the full target resolution while keys and values are spatially pooled by a factor $f$ to form a downsampled global receptive field. This preserves positional geometry close to the model’s native training resolution, enabling global attention over a downsampled representation at $1/f^2$ the attention cost, since a coarse view suffices to capture the low-frequency structure and overall composition formed in early denoising steps. Since the query side is never reduced, the output remains at full resolution and requires no upsampling. Specifically, each full-resolution query attends to the $N/f^2$ pooled keys and values and produces one output token, preserving the original $N$-token resolution. Phase~I starts maximally coarse and gradually decreases $f$ as denoising progresses and finer details begin to form. We describe the pooling strategy and this refinement schedule next.

To construct the downsampled key and value grids, AReA partitions the spatial grid into $f \times f$ blocks and selects one representative token from each block, using the block center by default (see Appendix~\ref{app:coarse-pool} for alternatives). Pooling alone reduces the number of keys but not their distances, since the pooled keys would keep their original full-resolution coordinates. AReA therefore places the pooled keys on a coarse $(W/f) \times (H/f)$ grid and maps the full-resolution query coordinates onto the same grid by scaling them by $1/f$, so that queries and keys use the same coordinate system. As a result, each $f \times f$ block of queries shares the position of its pooled key, keeping the relative distances encoded by RoPE close to the model's training range. It is worth noting that text tokens are never pooled, so each of the $N$ queries attends to all $T + N/f^2$ keys.

Moreover, the choice of $f$ involves a trade-off. A large $f$ keeps the coarse
grid at the model's native range but is too coarse to resolve details such as
object boundaries, while a small $f$ gives a finer grid beyond the native range,
causing the model to duplicate subjects (Figure~\ref{fig:ablation-ramp}). We
therefore introduce a \textit{pooling ramp}, which decreases $f$ linearly over
Phase~I following
\begin{equation}
\tilde f_t = f_0 + \frac{t-1}{S_c-1}\,\big(f_{\min}-f_0\big), \qquad t = 1,\dots,S_c,
\label{eq:factor}
\end{equation}
decaying the factor from an initial $f_0$ down to a floor of $f_{\min}=2$, with each value snapped to a valid grid divisor (Appendix~\ref{app:coarse-ramp}). As $f$ decreases, the coarse grid expands, and this gradual increase in resolution smoothly bridges the gap between the initial global composition and the fine details targeted in the next phase. An overview of this process is shown in Figure~\ref{fig:pipeline}.

\vspace{-3pt}
\subsection{Phase II: adaptive windowed fine attention}
\label{sec:method:phase2}

Once the global layout is established, the remaining steps focus on refining fine details, which Phase~I cannot provide, since pooling removes the local information needed at this stage. Fine details are inherently local, so at this stage, each token primarily depends on its surrounding neighborhood, making local-window attention sufficient. Windows are also where the speedups come from. Since attention scales quadratically with the number of tokens, processing multiple small windows is much cheaper than computing over the full grid. Moreover, as the window size decreases, the cost per window drops quadratically, while the number of windows grows only linearly, reducing the overall computational cost. However, windows are not always cheaper. A few large, overlapping windows, or simply too many windows, can cost more than a single full-attention step, so both the size and number of windows must be chosen carefully. The design goal is therefore to use the smallest windows that together remain cheaper than a single full-attention pass while providing enough local context. However, starting with small windows too early can fragment the image. Although Phase~I sets the global structure, neighboring regions must stay aligned during Phase~II. AReA therefore begins with large windows for coherence, then gradually shrinks them as details emerge, keeping within-window relative positions near the model’s training range.

Concretely, the window size $w_t$ changes at every denoising step. It starts at $w_{\text{start}}$, the largest window for which windowed attention is still cheaper than a full-attention pass, which is about $70\%$ of the longer side of the patch grid, and shrinks each step until it reaches the floor $w_{\min}$, which is the model's native range,
{
\begin{equation}
\label{eq:wsched}
w_t \;=\; w_{\text{start}} \Big(\tfrac{w_{\min}}{w_{\text{start}}}\Big)^{\frac{t-1}{S_d-1}}, \qquad t = 1, \dots, S_d,
\end{equation}
}
where $S_d$ is the number of steps in this phase. Both $w_{\text{start}}$ and subsequent window sizes are checked against a cost budget on image-token attention, so that no windowed step exceeds a full-attention pass in that cost (see Appendix \ref{app:window-cost}). At step $t$, the $H \times W$ grid is covered by a base tiling of $n_H \times n_W$ windows, each of size $w_t \times w_t$, where $n_H = \lceil H / w_t \rceil$ and $n_W = \lceil W / w_t \rceil$ are the minimum numbers of windows needed to cover each axis. The windows are spaced evenly so that the overlap is distributed equally between neighboring windows.
AReA then adapts this base tiling to the image content. A DWT map, computed from the current hidden states, measures the amount of fine detail in each region (see Appendix~\ref{app:dwt}). Windows over smooth, low-detail regions are split into four quadrants (visualized in \Cref{fig:pipeline}), where splitting is safe because a larger window over smooth content adds little information. In contrast, a window over a detailed region is left at its scheduled size, since splitting it would cut the structure it needs to see. This splitting saves computation, and we make sure that the adapted tiling does not cost more than the base tiling (see Appendix~\ref{app:adapt-cost}).

When the window size does not evenly divide the grid, adjacent windows overlap. Outputs for shared tokens are blended using center-weighted averaging, which favors windows where the token lies closer to the center. Despite this blending, windowed attention remains prone to visible seams because boundary tokens lack context from the other side. AReA addresses this in two ways. First, window boundaries are randomly shifted across steps, preventing seams from forming at fixed locations. Second, each window's keys and values are extended with a narrow ring of neighboring tokens ($\delta_{\max}$), which we call the \emph{Halo}, giving edge tokens access to context across the boundary (see Appendix~\ref{app:halo} for details). Since the queries and outputs remain unchanged, the additional cost is much smaller than that of enlarging the window itself. Even with the coherent global structure established in Phase~I, windowed attention can still duplicate small objects or local details, because each window cannot see beyond its boundaries and may reproduce content already present elsewhere in the image. AReA addresses this with a \emph{global-KV bridge}. At every Phase~II step, each window’s KV set is extended with the keys and values of a small set of tokens from the rest of the grid. Tokens are selected based on their deviation from the global mean key, prioritizing distinctive content that should be shared across windows, with the selection spatially distributed across the grid to ensure broad spatial coverage. The number of bridge tokens scales with the window as $R = (\rho w_t)^2$, maintaining a fixed fraction of each window’s keys ($\rho$) throughout the phase (see Appendix~\ref{app:gkv}).


Putting everything together, in Phase~II the queries of each window of side
$w_t$ (or $w_t/2$ if split) attend to its own local tokens, the $T$ text tokens,
and $E$ additional tokens from the Halo and the global-KV bridge. AReA keeps the RoPE frequencies unchanged and requires no RoPE length extrapolation. This phase therefore provides fine-grained local attention with
enough cross-window context to reduce seams and duplication.

%% file: sections/experiments.tex
\begin{figure}[t]
  \centering
  \includegraphics[width=\textwidth]{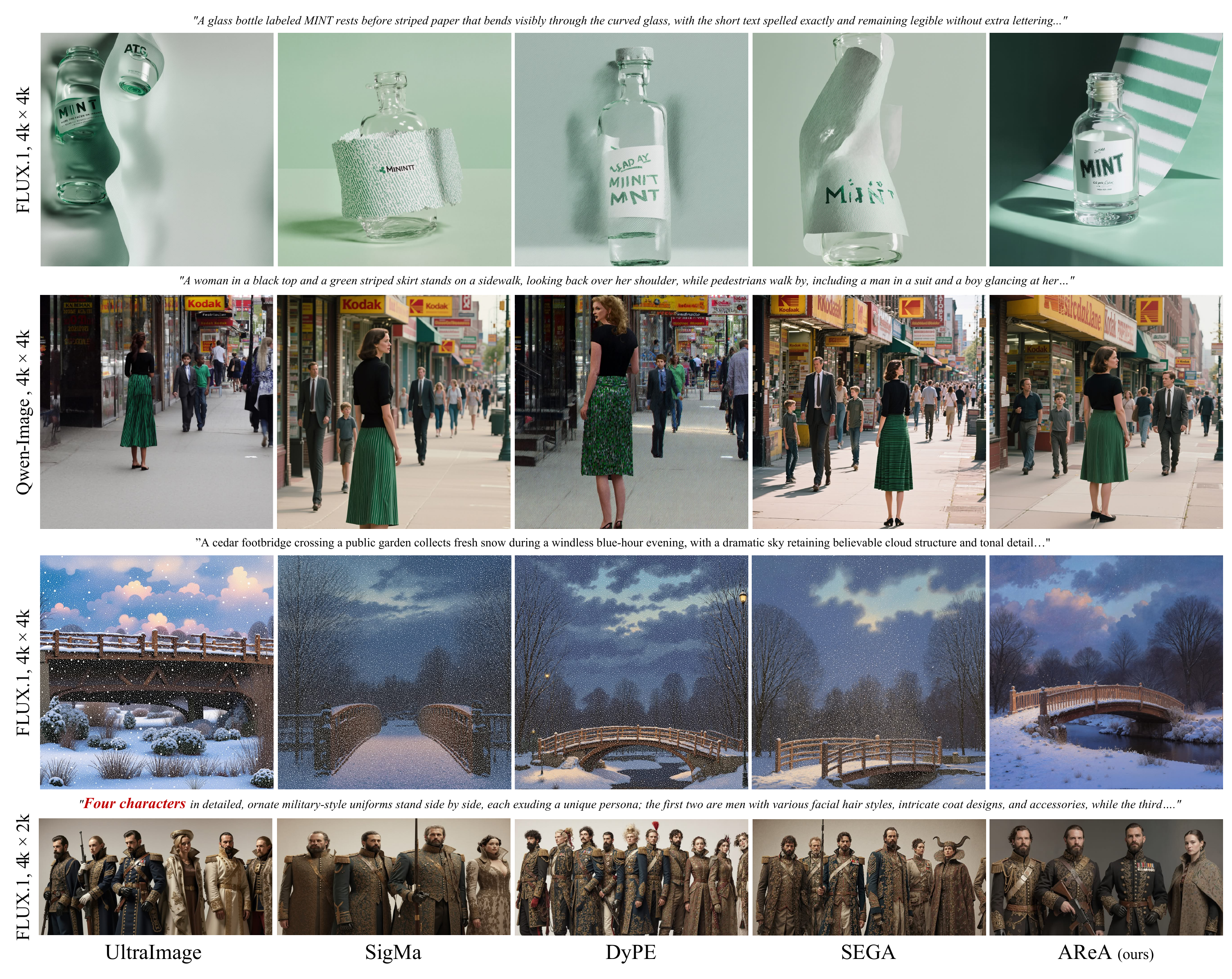}
  \vspace{-20pt}
  \caption{\textbf{Qualitative comparisons across resolutions.} AReA eliminates structural distortions and subject duplication seen in baseline methods, achieving superior fidelity and prompt alignment.
  \vspace{-10pt}
  }
  \label{fig:main-comparison}
\end{figure}

\vspace{-10pt}
\section{Experiments}
\label{sec:experiment}
\vspace{-5pt}

\paragraph{Setup.}
We evaluate AReA on FLUX.1~\citep{flux} and Qwen-Image~\citep{qwen}. Since these models differ in their noise schedules and default numbers of denoising steps, we select the switching threshold $\sigma^\star$ separately for each model. Based on an empirical analysis (Appendix~\ref{app:phase-switch}), we set $\sigma^\star$ such that the first 8 steps for FLUX.1 and the first 20 steps for Qwen-Image use cross-scale coarse attention, while the remaining steps use adaptive windowed fine attention. Since AReA is a direct-inference method that generates images end-to-end, we compare it with other direct-inference methods, including DyPE~\citep{dype}, UltraImage~\citep{ultraimage}, SigMa~\citep{sigma}, and SEGA~\citep{sega}. Comparisons with multi-stage methods on FLUX.1 are provided in Appendix~\ref{app:multistage}. All baselines are evaluated using their official implementations. Across all experiments, we set $\delta_{\max}$ to 8 for Halo and $\rho$ to 0.5 for global-KV bridge. All experiments are conducted on NVIDIA H100 GPUs.

\begin{wrapfigure}{r}{0.5\columnwidth}
  \centering
  \vspace{-5pt}
  \includegraphics[width=\linewidth]{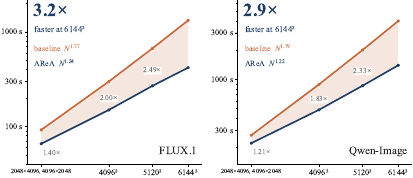}
  \vspace{-5pt}
  \caption{\textbf{Generation latency vs. resolution.} AReA is significantly faster than full-attention baselines across all resolutions, reaching $3.2\times$ and $2.9\times$ acceleration at $6144^2$ on FLUX.1 and Qwen-Image, respectively. \vspace{-5pt}}
  \label{fig:efficiency-plot}
\end{wrapfigure}

\paragraph{Evaluation.}

We evaluate on Aesthetic-4K~\citep{aesthetic4k}, which provides 195 prompts paired with high-resolution reference images. Inspired by SEGA~\citep{sega} and JoLT~\citep{jolt}, we further introduce \textbf{HRGen-1K}, a benchmark of 1,000 challenging and diverse LLM-generated prompts designed to rigorously evaluate high-resolution generation across complex scenes and compositions. Appendix~\ref{app:hrgen} describes its construction and statistics. We conduct quantitative experiments at four high-resolution settings with varying aspect ratios: $6144 \times 2048$, $2048 \times 4096$, $4096 \times 4096$, and $5120 \times 5120$. We evaluate image fidelity using FID~\citep{fid}, its patch-based variant FID$_p$ computed on local crops, and Inception Score (IS). Perceptual quality is assessed using Aesthetic Score~\citep{aesthetic} and the reference-free metrics MUSIQ~\citep{musiq}, CLIP-IQA~\citep{clipiqa}, and ARNIQA~\citep{arniqa}. Semantic alignment is measured by CLIP Score~\citep{clipscore}, while joint alignment and human-preferred visual quality are assessed using ImageReward (IR)~\citep{imagereward}, PickScore (PS)~\citep{pickscore}, and HPSv2~\citep{hps}. We also report latency as the average generation time per image in seconds.

\vspace{-5pt}
\subsection{State-of-the-art Comparison}

\paragraph{Qualitative comparison.}
As shown in Figure~\ref{fig:main-comparison}, existing direct-inference methods often distort the global structure of subjects at high resolutions, frequently producing stretched objects and bodies as well as struggling in generating texts. We attribute this to their RoPE modifications, which shift the model away from its training regime and disrupt spatial coherence across the image. In contrast, AReA preserves global structure and fine details while keeping relative positions within the training range, without RoPE length extrapolation. As a result, it also achieves stronger prompt alignment, particularly in composition and object counting. Further comparisons are in Appendix~\ref{app:qualitative}. 

\input{tables/main_results}

\input{tables/500_table}

\vspace{-5pt}
\paragraph{Quantitative comparison.}
As reported in Tables~\ref{tab:main-comparison} and~\ref{tab:flux-hrgen-1k}, AReA outperforms direct-inference baselines, including SEGA, DyPE, SigMa, and UltraImage, on most metrics for both FLUX.1 and Qwen-Image. Across all resolutions, including non-square aspect ratios and up to $5120 \times 5120$, it consistently ranks first or second in both semantic alignment and perceptual quality. While most competing methods degrade noticeably as the number of tokens grows, AReA maintains high fidelity and structural coherence. These gains also come with improved efficiency. As shown in Figure~\ref{fig:efficiency-plot} and Table~\ref{tab:main-comparison}, AReA achieves lower latency than all baselines at every resolution, with the advantage growing at larger resolutions where attention cost increasingly dominates.

\begin{figure}[t]
  \centering
  \includegraphics[width=\textwidth]{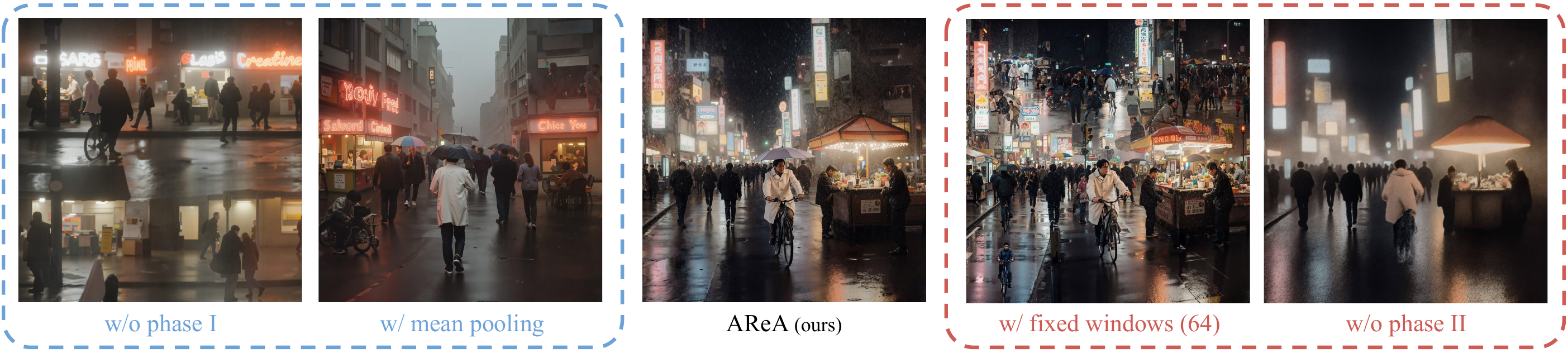}
  \vspace{-20pt}
  \caption{\textbf{Qualitative ablation on FLUX.1 at $\mathbf{4096 \times 4096}$.} Without Phase I, windows diverge and disrupt global layout. Without Phase II, the image lacks fine texture and remains blurry. Mean pooling degrades overall image quality (compared in Table~\ref{tab:flux_pooling_ablation}), while fixed windows without shrinking cause seams and content fragmentation compared to default AReA.
  \vspace{-10pt}
  }
  \label{fig:ablation}
\end{figure}

\input{tables/main_ablation}

\vspace{-5pt}
\subsection{Ablation Studies}
\label{sec:ablation}
\vspace{-5pt}

We conduct ablation studies at $4096 \times 4096$ on FLUX.1 using Aesthetic-4K,
first examining the contribution of each phase and then their design choices. As
shown in Table~\ref{tab:flux_ablation_4k} and Figure~\ref{fig:ablation}, without
Phase~I, the global structure is not established, and local windows evolve
independently, producing inconsistent content across the image. Without
Phase~II, the output lacks fine details despite a well-formed low-frequency
structure. Beyond improving efficiency, Phase~II limits how far and over how
many keys each token attends in later steps, avoiding diluted attention that
blurs fine details.

We further evaluate design choices within each phase. In Phase~I, keys and
values are pooled within blocks so that queries attend to a downsampled
representation of the image, with the initial scale factor $f$ keeping this
representation within the native resolution. As shown in
Table~\ref{tab:flux_ablation_4k}, starting with $f$ that yields a pooled
representation beyond the native resolution degrades performance, as queries
attend over distances unseen during training, pushing the model out of
distribution. Moreover, since denoising progresses from coarse to fine, queries
should attend to increasingly fine representations as sampling proceeds.
Keeping this coarse scale fixed throughout Phase~I passes only a weak
representation of the image to Phase~II, leaving a poorly formed global
structure on which to build fine details. Accordingly, we start from $f$ that
matches the native resolution and decrease it linearly in discrete steps during
Phase~I, equivalently increasing the resolution of the pooled keys and values.  Finally, as shown in Table~\ref{tab:flux_pooling_ablation}, selecting the block center as the representative token outperforms mean, random, and medoid pooling on nearly all metrics.


\input{tables/pooling_ablation}

Phase~II schedules the size, number, and positions of windows across denoising steps. As shown in Figure~\ref{fig:ablation}, using a fixed window size of 64, matching the native resolution of FLUX.1, breaks the image into inconsistent regions. Such outputs are unusable, yet they still score well on metrics that are insensitive to layout errors, such as Aesthetic and CLIP; the failure is instead reflected in lower ImageReward and PickScore and a higher FID. Similarly, removing the global-KV bridge leads to duplicated and repeated subjects across the image (Figure~\ref{fig:ablation-halogkv}), which leaves most metrics nearly unchanged but raises FID from 148.5 to 161.3. The Halo removes seams along window boundaries that are subtle and visible mainly in qualitative comparisons (Figure~\ref{fig:ablation-halogkv}).

%% file: tables/main_results.tex
\begin{table}[!t]
  \centering
  \caption{Comparison of AReA against state-of-the-art baselines on FLUX.1 and Qwen-Image across different resolutions and aspect ratios on Aesthetic-4K. \textbf{Bold}: best; \underline{underline}: second-best.}
  \label{tab:main-comparison}
  \vspace{0.4em}
  \setlength{\tabcolsep}{3pt}
  \renewcommand{\arraystretch}{1.05}
  \resizebox{\textwidth}{!}{%
    \begin{tabular}{clccccccc!{\color{black}\vrule}ccccccc}
    \toprule
      & & \multicolumn{7}{c!{\color{black}\vrule}}{\textbf{FLUX.1}}
      & \multicolumn{7}{c}{\textbf{Qwen-Image}} \\
    \cmidrule(lr){3-9}\cmidrule(lr){10-16}
      \textbf{Res.} & \textbf{Method}
& {Latency(s)$\downarrow$} & {ImageReward$\uparrow$} & {Aesthetic$\uparrow$} & {CLIP$\uparrow$} & {PickScore$\uparrow$} & {FID$\downarrow$} & {FID$_p\downarrow$}
& {Latency(s)$\downarrow$} & {ImageReward$\uparrow$} & {Aesthetic$\uparrow$} & {CLIP$\uparrow$} & {PickScore$\uparrow$} & {FID$\downarrow$} & {FID$_p\downarrow$} \\

    \midrule
    \multirow{6}{*}{\rotatebox[origin=c]{90}{2048$\times$4096}}
      & Base & 97 & -0.8696 & 5.91 & 26.49 & 20.41 & 171.4 & 67.4
                  & \underline{274} & -0.5186 & 5.96 & 27.25 & 20.83 & 167.6 & \textbf{57.8} \\
      & UltraImage & \underline{89} & -0.1474 & 6.25 & 28.11 & 21.28 & 173.3 & 79.0
                  & 301 & 0.4448 & 5.79 & 28.19 & 21.07 & 161.7 & \underline{62.6} \\
      & DyPE & 97 & 0.5927 & \underline{6.30} & 28.43 & 22.15 & 156.5 & 79.9
                  & \underline{274} & 0.5769 & 6.12 & 28.71 & 21.64 & 159.0 & 74.7 \\
      & SigMa & 103 & 0.6727 & 6.14 & 28.07 & 22.01 & \underline{152.5} & \underline{63.9}
                  & 275 & 0.6155 & 6.22 & 28.87 & 22.03 & 152.5 & 63.0 \\
      & SEGA & 98 & \underline{0.7863} & 6.28 & \underline{28.95} & \underline{22.47} & 153.9 & 76.1
                  & 276 & \underline{1.2467} & \underline{6.42} & \underline{29.89} & \underline{23.10} & \underline{146.1} & 69.6 \\
\rowcolor{gray!15}
      \cellcolor{white} & \textbf{AReA} & \textbf{66} & \textbf{1.0881} & \textbf{6.34} & \textbf{29.23} & \textbf{22.95} & \textbf{151.9} & \textbf{62.5}
                  & \textbf{221} & \textbf{1.3260} & \textbf{6.44} & \textbf{30.07} & \textbf{23.26} & \textbf{144.5} & 68.5 \\

\midrule
\multirow{6}{*}{\rotatebox[origin=c]{90}{6144$\times$2048}}
      & Base & 185 & -0.3800 & 5.82 & 25.15 & 20.73 & 173.5 & 83.6
                  & \underline{544} & 0.1886 & 5.86 & 26.92 & 21.10 & 175.3 & 68.2 \\
      & UltraImage & \underline{183} & 0.4103 & 6.11 & 27.55 & 21.81 & 171.4 & \underline{74.0}
                  & 552 & 0.3310 & 5.71 & 27.59 & 20.80 & 173.4 & \underline{63.1} \\
      & DyPE & 185 & \underline{0.8060} & \underline{6.24} & \underline{28.08} & \underline{22.34} & 173.3 & 97.4
                  & \underline{544} & 0.1442 & 5.60 & 26.60 & 20.47 & 181.7 & 100.6 \\
      & SigMa & 191 & 0.4321 & 5.88 & 26.69 & 21.43 & 181.3 & 86.1
                  & 545 & 0.6295 & 6.04 & 28.43 & 21.75 & 161.6 & 65.1 \\
      & SEGA & 188  & 0.5357 & 6.12 & 27.64 & 21.88 & \underline{168.2} & 94.2
                  & 548 & \textbf{1.3047} & \underline{6.33} & \textbf{29.28} & \textbf{23.22} & \underline{160.8} & 73.4 \\
\rowcolor{gray!15}
      \cellcolor{white} & \textbf{AReA} & \textbf{107} & \textbf{1.0450} & \textbf{6.26} & \textbf{29.02} & \textbf{22.74} & \textbf{161.0} & \textbf{62.0}
                  & \textbf{355} & \underline{1.2752} & \textbf{6.34} & \underline{29.16} & \underline{22.69} & \textbf{159.6} & \textbf{62.2} \\
    \midrule
    \multirow{6}{*}{\rotatebox[origin=c]{90}{4096$\times$4096}}
      & Base & 314 & -0.7177 & 5.80 & 25.72 & 20.34 & 179.8 & 112.0
                  & \underline{902} & -0.1313 & 5.95 & 27.20 & 20.84 & 167.9 & 77.2 \\
      & UltraImage & \underline{256} & -0.1379 & 6.11 & 26.77 & 20.85 & 169.4 & 82.0
                  & 983 & 0.6911 & 5.84 & 28.64 & 21.43 & 155.3 & 65.1 \\
      & DyPE & 314 & 1.1070 & \underline{6.32} & 28.98 & 22.79 & 155.2 & 95.4
                  & \underline{902} & 0.7669 & 6.05 & 28.70 & 21.62 & 165.1 & 82.1 \\
      & SigMa & 322 & 1.0149 & 6.16 & 28.36 & 22.47 & 156.8 & 83.3
                  & 906 & 1.2614 & 6.26 & 29.40 & 22.66 & 152.2 & \underline{60.8} \\
      & SEGA & 319 & \underline{1.2702} & 6.31 & \underline{29.16} & \underline{23.16} & \underline{150.6} & \underline{71.8}
                  & 906 & \underline{1.4548} & \underline{6.47} & \textbf{29.83} & \textbf{23.65} & \underline{147.4} & 65.6 \\
\rowcolor{gray!15}
      \cellcolor{white} & \textbf{AReA} & \textbf{151} & \textbf{1.3250} & \textbf{6.37} & \textbf{29.71} & \textbf{23.31} & \textbf{148.5} & \textbf{58.2}
                  & \textbf{487} & \textbf{1.4847} & \textbf{6.49} & \underline{29.62} & \underline{23.53} & \textbf{147.3} & \textbf{55.3} \\
    \midrule
    \multirow{6}{*}{\rotatebox[origin=c]{90}{5120$\times$5120}}
      & Base & 685 & -1.8620 & 5.15 & 19.32 & 18.68 & 252.7 & 212.5
                  & \underline{2014} & -0.6020 & 5.78 & 26.64 & 20.24 & 186.4 & 128.4 \\
      & UltraImage & \underline{546} & -1.2421 & 5.60 & 22.50 & 19.47 & 206.1 & 165.8
                  & 2057 & 0.2061 & 5.64 & 27.13 & 20.73 & 181.5 & 98.3 \\
      & DyPE & 685 & 0.8389 & 6.14 & 27.79 & 22.01 & 163.4 & 148.0
                  & \underline{2014} & -0.3734 & 5.38 & 25.06 & 19.86 & 202.8 & 178.9 \\
      & SigMa & 704 & 0.5278 & 5.88 & 26.01 & 21.43 & 178.5 & 225.2
                  & 2019 & 0.5555 & 5.90 & 27.83 & 21.38 & 172.5 & 100.5 \\
      & SEGA & 693 & \underline{1.1143} & \underline{6.25} & \underline{29.16} & \underline{22.86} & \underline{155.7} & \underline{85.0}
                  & 2020 & \underline{1.4192} & \underline{6.42} & \textbf{29.90} & \textbf{23.72} & \underline{149.2} & \underline{63.4} \\
\rowcolor{gray!15}
      \cellcolor{white} & \textbf{AReA} & \textbf{269} & \textbf{1.2319} & \textbf{6.29} & \textbf{29.34} & \textbf{22.94} & \textbf{152.1} & \textbf{63.3}
                  & \textbf{862} & \textbf{1.4540} & \textbf{6.50} & \underline{29.80} & \underline{23.27} & \textbf{146.7} & \textbf{55.8} \\
    \bottomrule
    \end{tabular}%
  }
  \vspace{0.3em}
\end{table}

%% file: tables/500_table.tex
\begin{wraptable}{r}{0.55\textwidth}
  \vspace{-15pt}
  \centering
  \scriptsize
  \caption{Comparison on FLUX.1 at $4096 \times 4096$ on HRGen-1K. \textbf{Bold}: best; \underline{underline}: second-best.}
  \vspace{5pt}
  \label{tab:flux-hrgen-1k}
  \setlength{\tabcolsep}{2.5pt}
  \renewcommand{\arraystretch}{1.05}
  \begin{tabular}{lcccccc}
    \toprule
    \textbf{Method} & IR$\uparrow$ & Aes.$\uparrow$ & CLIP$\uparrow$ & Pick$\uparrow$ & HPS$\uparrow$ & IS$\uparrow$ \\
    \midrule
    Base       & -1.4649 & 5.12 & 23.36 & 19.04 & 0.232 & 7.20 \\
    UltraImage & -0.0277 & 5.77 & 28.26 & 20.46 & 0.269 & 10.17 \\
    DyPE       & 0.3280 & \underline{5.83} & 29.35 & 20.94 & 0.266 & 10.93 \\
    SigMa      & 0.0627 & 5.63 & 27.91 & 20.54 & 0.258 & 10.70 \\
    SEGA       & \underline{0.4422} & 5.82 & \underline{29.63} & \underline{21.18} & \underline{0.270} & \underline{11.99} \\
    \rowcolor{gray!15}
    \textbf{AReA} & \textbf{0.6001} & \textbf{5.87} & \textbf{29.89} & \textbf{21.44} & \textbf{0.275} & \textbf{13.90} \\
    \bottomrule
  \end{tabular}
  \vspace{-1em}
\end{wraptable}

%% file: tables/main_ablation.tex
\begin{table}[t]
\centering
\scriptsize
\caption{Ablation study on FLUX.1 at $4096\times4096$ on Aesthetic-4K. Within each phase, the first row replaces the whole phase with full attention; the rows below change one component.}
\label{tab:flux_ablation_4k}
\setlength{\tabcolsep}{3pt}
\renewcommand{\arraystretch}{1.05}
\resizebox{\textwidth}{!}{%
\begin{tabular}{@{}cl cccccccc@{}}
\toprule
& Method & Latency(s)$\downarrow$ & ImageReward$\uparrow$ & Aesthetic$\uparrow$ & MUSIQ$\uparrow$ & PickScore$\uparrow$ & HPS$\uparrow$ & CLIP$\uparrow$ & FID$\downarrow$ \\
\midrule
\multirow{3}{*}{\rotatebox{90}{Phase I}}
& No Phase I (full attention) & 201 & -0.41 & 5.85 & 38.06 & 20.73 & 0.264 & 26.63 & 171.42 \\
& Flat $f = 2$ & 165 & 1.23 & 6.27 & \textbf{46.67} & 23.03 & 0.294 & 29.13 & 152.08 \\
& Flat $f = 4$ (no ramp) & 142 & 1.27 & \underline{6.38} & 44.72 & 23.17 & 0.292 & 29.63 & 154.21 \\
\midrule
\multirow{5}{*}{\rotatebox{90}{Phase II}}
& No Phase II (full attention) & 248 & 1.10 & 6.35 & 40.67 & 22.71 & 0.286 & 28.78 & 155.44 \\
& Fixed windows ($w = 64$) & 123 & 1.26 & \textbf{6.43} & 45.26 & 23.12 & \underline{0.296} & \textbf{29.78} & 152.15 \\
& w/o Global-KV bridge & 133 & \textbf{1.32} & 6.25 & \underline{45.38} & 23.25 & 0.294 & 29.11 & 161.25 \\
& w/o Halo & 145 & \textbf{1.32} & 6.34 & 45.28 & \underline{23.29} & \underline{0.296} & 29.64 & \underline{148.76} \\
& w/o DWT splits & 187 & \underline{1.31} & 6.35 & 45.06 & 23.22 & 0.285 & 29.53 & 149.63 \\
\midrule
\rowcolor{gray!15}
\multicolumn{2}{@{}l}{\textbf{AReA (full)}} & 151 & \textbf{1.32} & 6.37 & 45.22 & \textbf{23.31} & \textbf{0.297} & \underline{29.71} & \textbf{148.50} \\
\bottomrule
\end{tabular}}
\end{table}

%% file: tables/pooling_ablation.tex
\begin{wraptable}{r}{0.5\textwidth}
  \vspace{-1.2em}
  \centering
  \scriptsize
  \caption{Phase~I pooling strategy on FLUX.1 ($4096 \times 4096$, Aesthetic-4K). $^\dagger$Default.}
  \vspace{5pt}
  \label{tab:flux_pooling_ablation}
  \setlength{\tabcolsep}{2pt}
  \renewcommand{\arraystretch}{1.05}
  \begin{tabular}{@{}lccccccc@{}}
    \toprule
    Pooling & IR$\uparrow$ & Aes.$\uparrow$ & MUSIQ$\uparrow$ & Pick$\uparrow$ & HPS$\uparrow$ & CLIP$\uparrow$ & FID$\downarrow$ \\
    \midrule
    Mean   & 1.23 & 6.30 & 44.93 & 22.84 & 0.289 & 29.28 & 152.78 \\
    Random & 1.25 & \underline{6.35} & \textbf{45.59} & \underline{23.22} & \underline{0.293} & \underline{29.37} & 152.53 \\
    Medoid & \underline{1.26} & 6.31 & 44.04 & 23.19 & \underline{0.293} & 29.19 & \underline{151.82} \\
    \rowcolor{gray!15}
    Center$^\dagger$ & \textbf{1.32} & \textbf{6.37} & \underline{45.22} & \textbf{23.31} & \textbf{0.297} & \textbf{29.71} & \textbf{148.50} \\
    \bottomrule
  \end{tabular}
  \vspace{-1em}
\end{wraptable}

%% file: sections/conclusion.tex
\vspace{-5pt}
\section{Conclusion}
\label{sec:conclusion}
\vspace{-5pt}

We presented AReA, an adaptive receptive-field attention mechanism that enables
efficient and coherent end-to-end high-resolution generation with diffusion
models, without any training and across different DiT backbones. Our key observation is that attention becomes increasingly local during denoising, so the receptive field that attention needs changes over time. AReA follows this coarse-to-fine progression in two phases. In early steps, it attends globally over pooled keys and values at a scale close to the native resolution to form the global structure, and in later steps, it attends within local windows that shrink and adapt to image content to recover fine details. This keeps token interactions close to the training regime without RoPE extrapolation, improving global consistency and fine-grained detail at resolutions far beyond training while reducing attention cost. In addition to the spatial redundancy addressed here, consecutive denoising steps also exhibit temporal redundancy, which we leave to future work for further acceleration.

%% file: sections/appendix.tex
\section*{\Large Appendix}

\section{Phase I: pooling factor and frame extent}
\label{app:coarse}

In Phase I, we use full-resolution queries, and keys and values are pooled in $f \times f$ blocks with query coordinates divided by $f$, and steps cost $1/f^{2}$. This section covers the two choices it defers, which factor to run at on which step and which token represents a block.

Both the start and the end of this schedule depend on the size of the pooled grid. This grid is $(W/f_t) \times (H/f_t)$, so its longer side contains $\max(H,W)/f_t$ patches, which is the effective resolution the model operates at. At $4096^2$ on FLUX.1, for instance, the patch grid is $256 \times 256$ and the native training grid is $64 \times 64$. A factor of $4$ therefore matches the native size exactly, while a factor of $2$ yields a $128 \times 128$ grid, twice the native size. For non-square resolutions (e.g., $6144 \times 2048$), we start from the smallest factor that keeps both sides within the native size. The schedule thus starts at or below the native size and ends above it. Starting within the native size keeps the early steps in distribution, while ending above it gives Phase~II a finer representation of the image to build on.

\subsection{Why the factor ramps}
\label{app:coarse-ramp}

\begin{figure}[t]
  \centering
  \includegraphics[width=\textwidth]{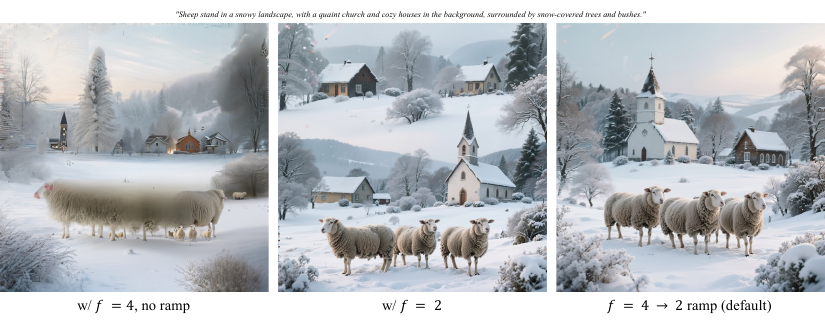}
  \vspace{-20pt}
  \caption{\textbf{Effect of the Phase~I pooling schedule on FLUX.1 at $4096 \times 4096$.} A fixed $f = 4$ forms the layout on a $64 \times 64$ pooled grid, which matches the training resolution but is too coarse to separate objects, so the sheep merge into a single blurred mass. A fixed $f = 2$ forms the layout on a $128 \times 128$ grid, twice the training resolution, producing scattered and repeated subjects. Decreasing $f$ from $4$ to $2$ first sets the layout at the training resolution and then refines it on the finer grid, giving separated subjects and a coherent scene.
  \vspace{-10pt}
  }
  \label{fig:ablation-ramp}
\end{figure}

Keeping $f_t = f_0$ (e.g., $f_0 = 4$ at $4096^2$) throughout Phase~I never exceeds the native size. The resulting layout is in distribution and globally consistent, but it is formed entirely on a $64 \times 64$ grid, which is too coarse to resolve finer structure such as object boundaries. Keeping $f_t = 2$ throughout fails in the opposite way. Every step then runs at twice the native size, so the layout forms at relative distances unseen during training, resulting in duplicated subjects and degraded global structure. In both cases, Phase~II cannot fix the problem, since its local attention only refines details within the layout set by Phase~I. Our schedule avoids both failures by starting at $f_0$ and decreasing $f_t$ as the layout forms, so the finer grids refine an existing layout rather than creating one. Figure~\ref{fig:ablation-ramp} shows both failure modes, with fused subjects lacking clear boundaries under a fixed $f = 4$ and scattered, duplicated content under a fixed $f = 2$.

Eq.~\ref{eq:factor} defines a target factor $\tilde f_t$ for each step, but block pooling also requires $f_t$ to divide both the height and width of the patch grid, so that the grid splits into whole $f_t \times f_t$ blocks. Rounding $\tilde f_t$ to the nearest integer does not guarantee this. We therefore set each $f_t$ to the common divisor of the grid height and width that is closest to $\tilde f_t$, restricted to values no larger than $f_{t-1}$ so that pooling never becomes coarser from one step to the next,
\begin{equation}
f_t = \operatorname*{arg\,min}_{\substack{d \in \mathcal{D},\ d \le f_{t-1}}} \big|\, d - \tilde f_t \,\big|,
\quad t = 2,\dots,S_c-1, \qquad f_1 = f_0,\ \ f_{S_c} = f_{\min},
\label{eq:fsnap}
\end{equation}
\begin{equation}
\mathcal{D} = \{\, d \in \mathbb{N} : d \mid \gcd(H, W) \,\}, \qquad
f_0 = \min\{\, d \in \mathcal{D},\ d \ge 2 : \max(H, W)/d \le P \,\},
\label{eq:f0}
\end{equation}
where $P$ is the native grid size in patches ($64$ for FLUX.1, and we use the same value for Qwen-Image). At $4096^2$, the grid size of $256$ has only $1$, $2$, and $4$ as divisors up to $f_0 = 4$. Since $f_{\min} = 2$, the schedule takes only two values, $[4,4,4,4,2,2,2,2]$, with four steps at a pooled grid size of $64$ followed by four at $128$. Resolutions whose grid has more divisors yield a more gradual schedule. At $6144^2$, for example, the schedule is $[6,6,4,4,4,3,3,2]$, so the pooled grid size grows from $64$ to $96$, $128$, and $192$. We set $f_{\min} = 2$ rather than $1$, since $f = 1$ applies no pooling and would turn the final steps of Phase~I into full attention at the target resolution, losing the efficiency gain in those steps.

Each Phase~I step costs $1/f_t^2$ of a full-attention pass. At $4096^2$, the eight Phase~I steps would cost $8$ passes with full attention. A fixed $f_t = 4$ reduces this to $0.50$, a fixed $f_t = 2$ to $2.00$, and the ramp to $1.25$. The ramp is therefore more expensive than a fixed $f_t = 4$, which never refines the layout, but cheaper than a fixed $f_t = 2$, which uses the finer grid from the first step.

\begin{figure}[t]
  \centering
  \includegraphics[width=\textwidth]{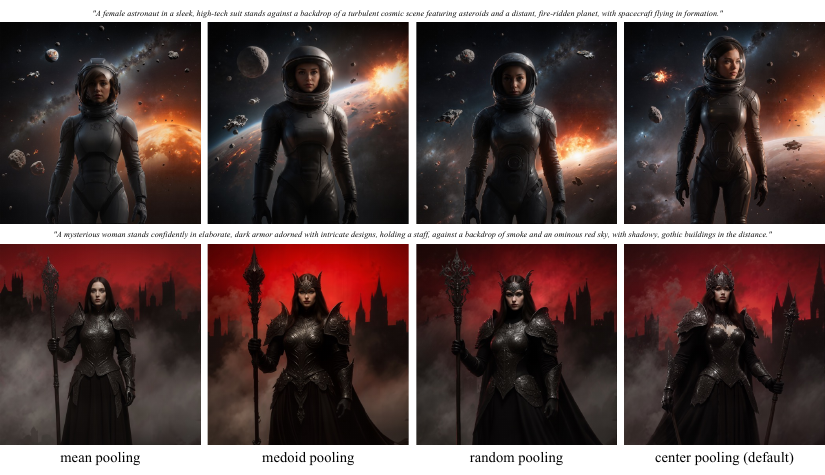}
  \vspace{-20pt}
  \caption{\textbf{Effect of the Phase~I pooling strategy on FLUX.1 at $4096 \times 4096$.} Mean pooling averages away the components that distinguish keys within a block, blurring fine detail such as armor ornamentation and facial features. Medoid selection keeps a real token but favors the most typical one in each block, partially inheriting this loss. Random selection preserves more detail but depends on the seed. Center pooling (default) uses a fixed, content-free token per block and gives the sharpest and most coherent results.
  \vspace{-10pt}
  }
  \label{fig:ablation-pooling}
\end{figure}

\subsection{Which token represents a block}
\label{app:coarse-pool}

Phase~I selects an actual token from each block rather than averaging the block. Averaging may seem like the natural choice, but it performs worst on most metrics among the four variants in Table~\ref{tab:flux_pooling_ablation}. Keys share a substantial common component across tokens, so averaging the $f^2$ keys in a block preserves this shared part while shrinking the token-specific components by up to a factor of $f$. As a result, the pooled keys from different blocks point in nearly the same direction, so a query assigns them similar attention weights and can no longer distinguish one block from another, which degrades overall image quality (Figure~\ref{fig:ablation-pooling}). Selecting an actual token instead preserves the original magnitude and direction of its key, and since its key and value come from the same token, the key--value pair remains consistent.

The second question is which token to select. A content-based rule, such as choosing the token closest to the block mean (medoid), favors the most typical token in the block and discards the distinctive ones, so it partially inherits the detail loss of mean pooling. Content-free rules avoid this, since random selection depends only on a random seed and center selection only on position within the block, so neither depends on the block's content. Both outperform medoid on most metrics (Table~\ref{tab:flux_pooling_ablation}, Figure~\ref{fig:ablation-pooling}).

Between the two content-free rules, the block center performs better on most metrics (Table~\ref{tab:flux_pooling_ablation}). Since the center is a fixed position, the same token is selected from each block at every step with the same $f$, so the other tokens in the block are not used as keys or values during these steps. Decreasing $f$ changes the blocks and therefore the selected tokens.

\begin{figure}[t]
  \centering
  \includegraphics[width=\textwidth]{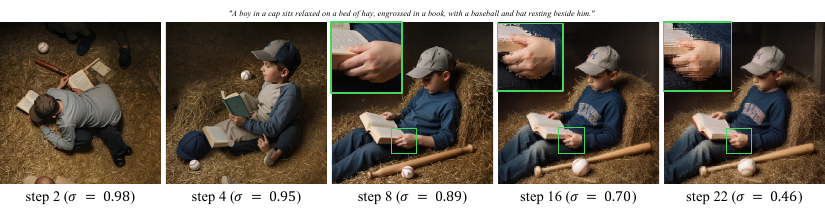}
  \vspace{-20pt}
  \caption{\textbf{Effect of switching phases too early or too late.} An early switch starts Phase~II before the structure has settled, degrading the final result. A late switch leaves too few Phase~II steps to add fine details, so the final image lacks detail.
  \vspace{-10pt}
  }
  \label{fig:sigma-ablation}
\end{figure}

\input{tables/sigma-ablation}

\subsection{When to switch phases}
\label{app:phase-switch}

To study where to switch between the two phases, we evaluate different values of $\sigma^\star$ on FLUX.1 by varying the denoising step at which AReA moves from Phase~I to Phase~II. Quantitative results are reported in Table~\ref{tab:sigma-ablation}, and qualitative comparisons are shown in Figure~\ref{fig:sigma-ablation}. Switching too early ends Phase~I before the global structure has settled, and the local windows of Phase~II cannot form it on their own. Switching too late has the opposite problem, since Phase~I continues into steps that should refine details, leaving too few Phase~II steps to add them and resulting in artifacts. Between these two extremes, results remain stable over a range of $\sigma^\star$, with $\sigma^\star = 0.89$ giving the most robust results on FLUX.1. We choose the default $\sigma^\star$ for the other backbones with the same procedure, giving $0.86$ for Qwen-Image, $0.94$ for FLUX.2, and $0.91$ for Ideogram~4, all of which fall within the stable range given their different noise schedules, timestep distributions, and training strategies.

\section{Phase~II: Window Attention Cost and Schedule}
\label{app:window-cost}

This section describes the cost of window attention in Phase~II. Let $H$ and $W$ denote the height and width of the patch grid, so the image has $HW$ tokens. A window of side $w$ contains $n = w^2$ image tokens. Since each window also attends to the $T$ text tokens, attention within a window operates on a joint sequence of $n + T$ tokens, and its cost grows quadratically with this length. We use this cost in two places. In this section, we use it to choose the starting window size and how the window size decreases over the Phase~II steps. In Appendix~\ref{app:dwt}, we use it to estimate the cost of the window splits made by the DWT-based adaptive tiling.

\subsection{Cost of a Windowed Step}
\label{app:step-cost-sec}

Each Phase~II step covers the image with the base tiling of Section~\ref{sec:method:phase2}, which uses the fewest windows of side $w$ needed to cover each axis,
\begin{equation}
\label{eq:n-axis}
N(w) = \Bigl\lceil \frac{H}{w} \Bigr\rceil \times \Bigl\lceil \frac{W}{w} \Bigr\rceil,
\end{equation}
which equals $n_H \times n_W$ at $w = w_t$. When $w$ does not divide an axis exactly, neighboring windows overlap slightly, and this overlap is spread along the axis.Randomly shifting the boundaries between interior windows changes where the windows are placed, but not their number or size, so it does not change the cost. For simplicity, we count only image tokens and assume each window attends over its own $w^2$ tokens. A windowed step then computes $N(w) \times w^4$ query-key products, while one full-attention pass over the image computes $(HW)^2$. We therefore measure the cost of a windowed step relative to one full-attention pass,
\begin{equation}
\label{eq:step-cost}
C(w) = N(w) \times \frac{w^4}{(HW)^2},
\end{equation}
so that full attention has $C = 1$. Ignoring the overlap introduced by the ceilings, $N(w) \approx HW/w^2$, so a step computes about $HW.\,w^2$ products and $C(w) \approx w^2/(HW)$. As $w$ decreases, the cost of each window falls as $w^4$ while the number of windows grows only as $w^{-2}$, so halving the window size reduces the cost of a step by a factor of four. Section~\ref{app:cost-lattice} discusses the window sizes at which the ceilings make this approximation inaccurate.

Eq.~\ref{eq:step-cost} counts only the $w^2$ image tokens in each window. In practice, each window runs a single joint attention that also includes the $T$ text tokens and $E$ extra keys (the halo ring and the global-KV bridge). The queries are the $T$ text tokens and the window's $w^2$ image tokens, while the keys are the same tokens plus the $E$ extra keys, which never act as queries. Each window therefore computes $(w^2 + T)(w^2 + T + E)$ query-key products rather than $w^4$. Since every window produces its own output for each text token, these outputs are averaged across windows. Keeping the text tokens as queries in every window adds cost, but removing them causes object artifacts and weaker prompt alignment, while reducing generation time by only $2.5\%$. When choosing the window schedule, we account for the extra keys by multiplying Eq.~\ref{eq:step-cost} by $(1 + E/w^2)$, and we omit the text tokens, since their relative cost stays modest for the window sizes in the base schedule and changes slowly with $w$, so omitting them barely affects the comparison between window sizes. We include them again in Appendix~\ref{app:dwt}, since each split adds windows that each process all $T$ text tokens as queries, so the text cost grows with the number of windows.

\subsection{Why Smaller Windows Are Not Always Cheaper}
\label{app:cost-lattice}

One might expect smaller windows to always be cheaper, but the cost in Eq.~\ref{eq:step-cost} does not always decrease as $w$ decreases. The reason is that the number of windows $N(w)$ is a step function. When $w$ decreases slightly, the same number of windows may no longer cover an axis, so a whole extra row and column of windows is needed. The cost of these extra windows can outweigh the savings from making each window smaller.

For example, on a $256 \times 256$ grid, $w = 128$ covers each axis with exactly two windows and no overlap, giving four windows in total and a cost of $C = 0.25$, that is, a quarter of a full-attention pass. Reducing the window by one to $w = 127$ means two windows no longer cover an axis, so three are needed along each axis. Together, these three windows cover $381$ tokens along an axis that has only $256$, so neighboring windows overlap by about $62$ tokens, roughly half their width. The resulting nine windows process about $2.2$ times as many tokens as the image contains and cost $C = 0.545$, more than twice as much as $w = 128$, even though each window is smaller.

The effect is strongest for large windows. Every size from $w = 129$ to $w = 255$ needs two windows along each axis, so each step uses four windows, each costing $(w/256)^4$ of a full-attention pass. Once each window costs more than a quarter of a full pass, which happens for $w \ge 182$, the four windows together cost more than full attention, reaching $C = 3.94$ at $w = 255$. Above $w = 181$, the only size on a square grid that is not more expensive than full attention is $w = \max(H, W)$, a single window covering the whole image, which gives $C = 1$. On a rectangular grid, even this choice exceeds full attention, since a single window of side $\max(H, W)$ costs $\max(H, W)^4/(HW)^2 > 1$.

Smaller windows are therefore not always cheaper, and some window sizes cost more than full attention. To keep every Phase~II step cheaper than full attention, the starting window size and the window schedule must account for these jumps. The next two subsections describe how we choose them.

\subsection{Choosing the Start Size}
\label{app:start-size}

Using the step cost $C(w)$ from Eq.~\ref{eq:step-cost}, Phase~II starts with the largest window for which a windowed step is still cheaper than a full-attention pass,
\begin{align}
w_{\text{start}} &=
\begin{cases}
\max\{\, w \ge w_h : C(w) < 1,\ N(w) > 1 \,\} & \text{if this set is non-empty},\\
w_h & \text{otherwise},
\end{cases}
\label{eq:wstart}\\[4pt]
w_h &= \max\!\big(w_{\min},\ \lfloor \max(H,W)/f_{S_c} \rfloor\big),
\label{eq:start-size}
\end{align}
where $f_{S_c}$ is the pooling factor at the last step of Phase~I, so $\max(H,W)/f_{S_c}$ is the longer side of the pooled grid at that step. The lower bound $w_h$ therefore keeps the first windows at least as large as this side and never smaller than $w_{\min}$. The condition $C(w) < 1$ ensures that, under the cost model of Eq.~\ref{eq:step-cost}, a windowed step is cheaper than full attention, which is the purpose of using windows. The condition $N(w) > 1$ makes explicit that a single window covering the whole image is excluded, since that case is simply full attention.

On a $256 \times 256$ grid, this rule gives $w_{\text{start}} = 181$. Every window size from $182$ to $255$ costs more than a full-attention pass. At this size, the image is tiled by two windows per axis, each covering just over $70\%$ of the axis and about half of the image by area. The same rule gives $226$ on a $320 \times 320$ grid and $271$ on a $384 \times 384$ grid. The first Phase~II step is therefore the most expensive one, and the shrinking schedule in Section~\ref{app:ladder} keeps every later step cheaper than the first. On square grids, $w_{\text{start}}$ is about $70\%$ of the side, since four windows of side $w$ cost less than full attention only when $w < S/\sqrt{2}$ for a grid of side $S$. If no window size qualifies, Phase~II starts at the lower bound $w_h$ of Eq.~\ref{eq:start-size}, so its first windows are at least as large as the pooled grid that Phase~I last used. This happens, for example, on strongly non-square grids (e.g., $384 \times 128$ at $6144 \times 2048$), where windows are clipped to the shorter side and Eq.~\ref{eq:step-cost} overestimates their cost.

\subsection{The Window Size Schedule}
\label{app:ladder}

Over the $S_d$ steps of Phase~II, the window shrinks from $w_{\text{start}}$ to the floor $w_{\min}$, the model's native training size in patches ($64$ for FLUX.1). Eq.~\ref{eq:wsched} defines this shrinkage over real values as a geometric schedule, in which the window is reduced by the same ratio at every step, so it shrinks quickly at first and more slowly as it approaches $w_{\min}$. Since window sizes must be integers, the schedule used in practice is its rounded form,
\begin{equation}
\label{eq:ladder-ideal}
\tilde w_{t}
\;=\;
\max\Bigl(w_{\min},\;
\Bigl\lfloor w_{\mathrm{start}}
\Bigl(\tfrac{w_{\min}}{w_{\mathrm{start}}}\Bigr)^{\frac{t-1}{S_d-1}}
\Bigr\rceil\Bigr),
\qquad t = 1,\dots,S_d,
\end{equation}
where $\lfloor \cdot \rceil$ denotes rounding to the nearest integer. The formula gives $\tilde w_1 = w_{\text{start}}$ and $\tilde w_{S_d} = w_{\min}$ exactly, so the schedule always begins at the start size and ends at the floor.

Rounding alone is not sufficient, for two reasons. First, the number of windows in Eq.~\ref{eq:n-axis} changes in discrete jumps because of the ceilings, so $C(w)$ can increase even when $w$ decreases, and a rounded size may land just past such a jump. Second, when consecutive values of the geometric schedule differ by less than one, rounding can produce the same size at two consecutive steps. We therefore adjust each rounded size by moving it down, never up, to the largest size that satisfies three conditions,
\begin{equation}
\label{eq:ladder-capped}
w_{t}
\;=\;
\max\bigl\{\,w \le \tilde w_{t} \;:\;
C(w)\le C(w_{\mathrm{start}}),\;
w \le w_{t-1}-1,\;
w \ge w_{\min} + (S_d - t)\,\bigr\},
\qquad t \ge 2,
\end{equation}
with $w_1 = w_{\text{start}}$. The first condition ensures that no step costs more than the first one, so the cost of a Phase~II step never rises above its starting point. The second requires consecutive sizes to strictly decrease, so the window changes at every step rather than repeating a size. The third requires step $t$ to leave room for the $S_d - t$ steps that remain, so that every later step can still take a distinct, smaller size and the last step lands exactly on $w_{\min}$. If no size satisfies all three conditions at some step, we use the cheapest admissible size instead. If the range $[w_{\min}, w_{\text{start}}]$ contains fewer than $S_d$ integers, it cannot hold $S_d$ distinct sizes at all, so strictly decreasing sizes are impossible and we use Eq.~\ref{eq:ladder-ideal} unchanged.

\begin{wrapfigure}{r}{0.5\columnwidth}
  \centering
  \vspace{-15pt}
  \includegraphics[width=\linewidth]{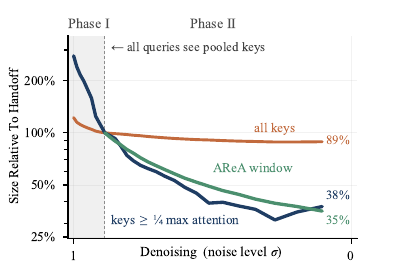}
  \vspace{-15pt}
  \caption{\textbf{Window size follows the locality of the model's attention.} AReA's window size at $4096^2$ and the model's attention range at its native resolution of $1024^2$, computed over keys receiving at least a quarter of the maximum attention (blue) and over all keys (orange). All curves are shown relative to their values at the start of Phase~II. In Phase~I, every query attends to pooled keys from the whole image, so no window is shown. In Phase~II, the window size follows a similar trend to the strongest attention, while attention over all keys remains broad. \vspace{-10pt}}
  \label{fig:windows-schedule-trend}
\end{wrapfigure}

The cost condition in Eq.~\ref{eq:ladder-capped} is still needed after choosing $w_{\text{start}}$ with Eq.~\ref{eq:wstart}, because a smaller window can cost more than a larger one later in the schedule. With the default start, the rounded schedule happens to satisfy this condition at every step, so Eq.~\ref{eq:ladder-capped} returns Eq.~\ref{eq:ladder-ideal} unchanged. The condition becomes active when the start lies just above a point where the number of windows increases. Starting at $w = 130$, for instance, the next rounded size of $125$ costs $0.51$, so the condition moves it down to $106$. It is therefore what guarantees that no step costs more than the first, for any start.

At $4096^2$ ($H = W = 256$, $S_d = 20$, \\ $w_{\min} = 64$), the schedule is
\[
\begin{aligned}
& 181,\,171,\,162,\,154,\,145,\,138,\,130,\,123,\,117, \\ 
& 111,\, 105,\,99,\,94,\,89,\,84,\,80,\,75,\,71,\,68,\,64,
\end{aligned}
\]
The window side shrinks by $10$ tokens in the first step and by only $3$ to $4$ tokens near the end. The window side shrinks by $10$ tokens in the first step and by only $3$ to $4$ tokens near the end. Under the image-only cost of Eq.~\ref{eq:step-cost}, Phase~II costs $6.61$ full-attention passes in total, compared with $20$ for full attention over the same steps, and no single step costs more than one pass. As shown in Figure~\ref{fig:windows-schedule-trend}, the window size decreases over denoising following a similar trend to the locality of attention, where each query's most-attended keys lie increasingly close to it as denoising proceeds (both attention curves are taken from Figure~\ref{fig:locality}).

\subsection{Deduplication and Coverage}
\label{app:dwt-dedupe}

Neighboring windows in the base tiling often overlap, so when two neighbors are both split, their facing quadrants can cover nearly the same region. Before attention is computed, we therefore merge pairs of quadrants whose intersection covers at least $80\%$ of the smaller quadrant, replacing each pair with its bounding box. Pairs are merged greedily, starting with the largest overlap. This removes a nearly redundant attention call and prevents the tokens in these regions from being counted twice when the outputs of overlapping windows are averaged. After splitting and merging, every image token is still covered by at least one window, and the cost of the step stays within the budget $B$ of the base tiling.

\section{Content-Adaptive Window Tiling}
\label{app:dwt}

The schedule in Appendix~\ref{app:ladder} sets one window size per step, and Eq.~\ref{eq:n-axis} turns this size into a base tiling of $N(w_t)$ windows. AReA then adjusts this tiling using a DWT map computed from the current hidden states. Since the map is recomputed at every Phase~II step, the tiling follows the image as its details emerge. The adaptation can change how computation is distributed across the image, but its total cost never exceeds that of the base tiling.

\subsection{The DWT Map}
\label{app:dwt-map}

Let $x \in \mathbb{R}^{B \times H \times W \times D}$ denote the image tokens of the current hidden states, reshaped onto the $H \times W$ patch grid. We first reduce them to a single value per patch by taking the standard deviation across channels and averaging over the batch,
\begin{equation}
\label{eq:scalar-map}
s_{r,c}
\;=\;
\frac{1}{B}\sum_{b=1}^{B}\mathrm{std}_{d}\bigl(x_{b,r,c,d}\bigr),
\end{equation}
and then subtract the spatial mean of $s$. The channel-wise standard deviation summarizes the features of each token in one value that changes sharply across edges and textured regions, which makes it a cheap input for the transform. The channel mean is less suitable, since changes of opposite sign in different channels cancel out when averaged, which can hide edges before the transform sees them.

We then apply a separable Haar transform for $L = 3$ levels. For each $2 \times 2$ block of values, with $p_1, p_2$ in the top row and $p_3, p_4$ in the bottom row, the low-pass band and the three detail bands are
\begin{align}
\label{eq:haar}
A   &= \tfrac{1}{4}(p_1+p_2+p_3+p_4), \nonumber\\
D_{v} &= \tfrac{1}{4}(p_1+p_2-p_3-p_4), \qquad
D_{h}  = \tfrac{1}{4}(p_1-p_2+p_3-p_4), \qquad
D_{d}  = \tfrac{1}{4}(p_1-p_2-p_3+p_4),
\end{align}
and the detail energy of the block is $|D_{v}|+|D_{h}|+|D_{d}|$. The next level applies the same transform to $A$. Each level's detail energy is upsampled back to the full grid and added to the map with weight $2^{-\ell}$, where $\ell = 1$ is the finest level. A coefficient at level $\ell$ covers a $2^{\ell} \times 2^{\ell}$ square of patches with a single value, so coarser levels produce large uniform regions. With equal weights, these regions would dominate the map and spread high values into the smooth areas around an edge, so the weights favor the finer levels, which localize edges more precisely. Finally, the map is divided by its maximum, giving the normalized map $E_{r,c}$. Each window $k$ of the base tiling is then scored by $e_k = \mathrm{mean}(E|_k)$, the mean of $E$ over the patches it covers.

\begin{figure}[pt]
  \centering
  \includegraphics[width=0.9\textwidth]{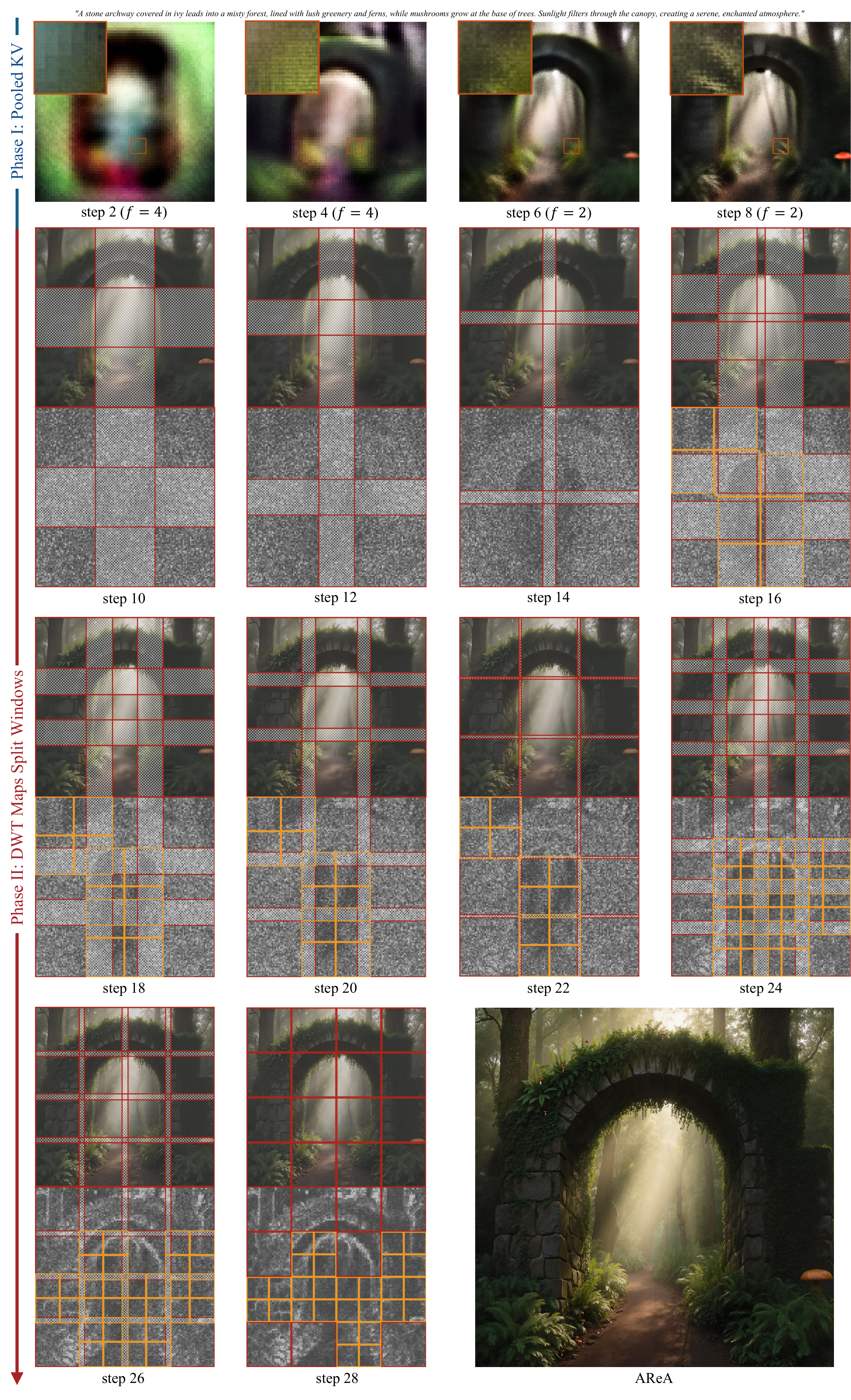}
  \caption{\textbf{AReA Across Denoising on FLUX.1 at 4096 × 4096}. \emph{Top row}: Phase~I steps with pooled keys and values (insets show the pixelation effect). \emph{Other rows}: Phase~II steps, each showing the current image (upper half) and its DWT detail map (lower half), with base windows in red, overlap regions hatched, and DWT splits in orange.
  \vspace{-10pt}
  }
  \label{fig:dwt-steps}
\end{figure}

\subsection{When Adaptation Is Applied}
\label{app:dwt-gate}

The windows to adapt are chosen by comparing their scores $\{e_k\}$ with one another, which always marks some windows as high and others as low, even when $E$ is essentially uniform noise. We therefore apply adaptation only when $E$ varies enough across the image, measured by the coefficient of variation of $\bar E$, the map $E$ averaged over tiles of side $w_t/2$,
\begin{equation}
\label{eq:gate}
u \;=\; \frac{\mathrm{std}(\bar E)}{\mathrm{mean}(\bar E)},
\qquad
\alpha \;=\;
\mathrm{clip}\!\left(\frac{u-u_{\mathrm{lo}}}{u_{\mathrm{hi}}-u_{\mathrm{lo}}},\,0,\,1\right),
\qquad
u_{\mathrm{lo}}=0.05,\; u_{\mathrm{hi}}=0.25.
\end{equation}
When $\alpha = 0$, the base tiling is used unchanged. Since the tile size follows $w_t$, the large tiles early in Phase~II average out fine-scale noise and keep the gate closed, while the small tiles near the floor respond to real structure. Each step can adapt at most $\lfloor \alpha\,N(w_t) \rfloor$ of its windows. Early steps with few windows are thus adapted only when the map is strongly non-uniform, and later steps with many windows are adapted the most. Figure~\ref{fig:dwt-steps} shows that the base tiling stays unchanged while the windows are still large (steps 10--14), and splits become more frequent as the windows approach the floor.

\subsection{Window Splitting and Its Cost}
\label{app:adapt-cost}

The base tiling remains the anchor, and each selected window is split once into four quadrants of half its side length. Quadrants smaller than $w_{\min}/2$ are not allowed, and quadrants are not split again within a step.

Since every window processes the $T$ text tokens as both queries and keys, a window $k$ with $n_k$ image tokens is charged
\begin{equation}
\label{eq:leaf-cost}
c(k) \;=\; \bigl(T + n_k\bigr)^2.
\end{equation}
A split is accepted only if the total cost stays within the budget of the base tiling, $B = \sum_{k \in \mathcal{U}_t} c(k)$. Splitting a window with $n$ image tokens changes the cost by
\begin{equation}
\label{eq:split-delta}
\Delta_{\mathrm{split}}
\;=\;
4\bigl(T+\tfrac{n}{4}\bigr)^2 - \bigl(T+n\bigr)^2
\;=\;
3\,T^{2} - \tfrac{3}{4}\,n^{2},
\end{equation}
since the image-image block shrinks from $n^2$ to $n^2/4$, while the text-text block is computed once per window and grows from $T^2$ to $4T^2$. A split is therefore cheaper only when $n > 2T$. On our default schedules, every window has at least $64^2 = 4096$ image tokens, far above $2T$ ($1024$ for FLUX.1), so every split is cheaper and $B$ acts as a safeguard. This cost omits the halo ring and the bridge registers, and since each quadrant adds its own halo and receives the full set of registers, the actual saving is smaller than $\tfrac{3}{4}n^2$.

A window is a candidate for splitting if its score $e_k$ is in the lowest half of all scores ($\tau = 0.5$) and strictly below the mean, where the second condition prevents ties on sparse maps from selecting windows. We split the smoothest windows because a smooth region gains little from a large receptive field and rarely shows a seam when divided, while splitting it still saves three quarters of its image-image cost. Windows over detailed regions are not split, so they keep the scheduled size $w_t$ and the larger context it provides.

\section{A Ring of Neighboring Keys (Halo)}
\label{app:halo}

\begin{figure}[t]
  \centering
  \includegraphics[width=\textwidth]{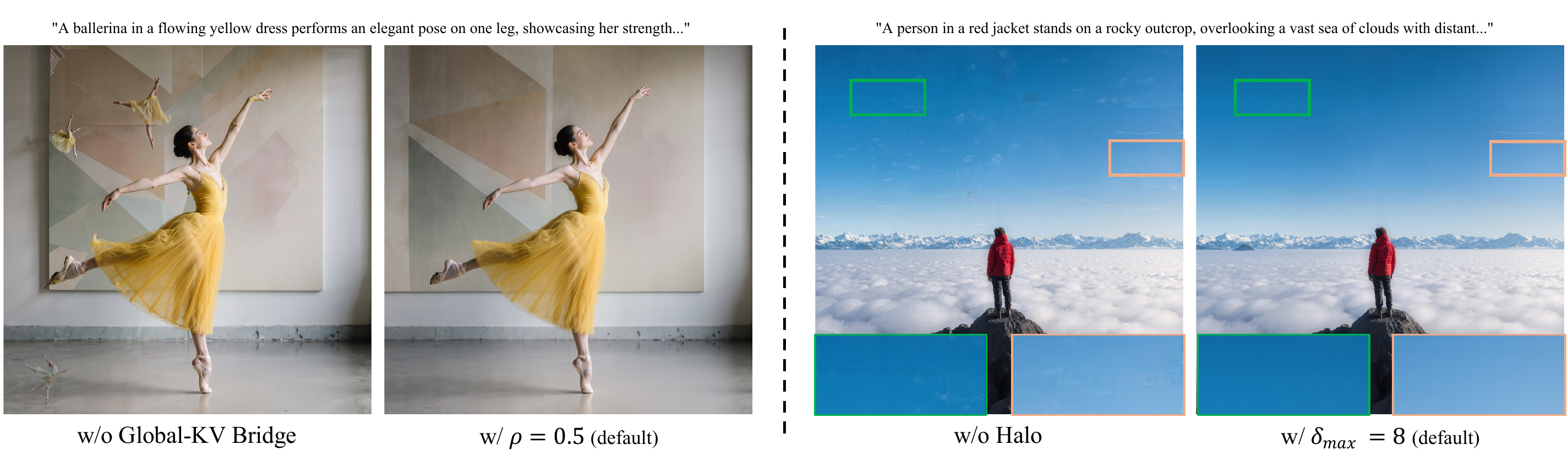}
  \vspace{-20pt}
  \caption{\textbf{Effect of the Global-KV Bridge and Halo}. \emph{Left}: without the global-KV bridge, each window is blind to content elsewhere on the canvas and paints extra copies of the subject. With the bridge, windows see distinctive tokens from the rest of the grid and the duplicates disappear. \emph{Right}: without the Halo, tokens at window edges lack context from across the boundary, leaving seams in smooth regions such as the sky.
  \vspace{-10pt}
  }
  \label{fig:ablation-halogkv}
\end{figure}

With windowed attention, a token at the edge of its window receives no context from the other side of the boundary, which is why seams remain visible even after the center-weighted merge. The halo addresses this locally. At every Phase~II step, the keys of each window are extended with the band of image tokens immediately surrounding it (Figure~\ref{fig:ablation-halogkv}, right).

For a window of side $w$ and a ring of width $\delta$, the band is the window grown by $\delta$ on every side, minus the window itself and clipped at the image boundary, so it adds up to $(w+2\delta)^2 - w^2$ keys. Only the keys and values grow. The queries, the output tokens, and the coverage counts used by the merge are unchanged, so the merge weights are identical to a run without the ring, and only the attention output changes. This is what distinguishes the halo from simply enlarging the window. An enlarged window would also use its extra tokens as queries, producing outputs for them and changing the coverage, at a cost of $(w+2\delta)^4$, whereas the halo costs only $w^2(w+2\delta)^2$.

The ring width is not fixed. At every step, each window draws $\delta$ uniformly from $\{0, \dots, \delta_{\max}\}$, with $\delta_{\max} = 8$ in all our runs. A ring of constant width would only move each window's boundary outward by the same amount, leaving the break in the same place at every step. Drawing the width randomly places the break differently for each window and each step. The draw depends only on the run seed and the step index, so for models with classifier-free guidance, the conditional and unconditional passes use the same band.

In the units of Eq.~\ref{eq:step-cost}, the ring multiplies the cost of a step by $\bigl((w+2\delta)/w\bigr)^2$, and together with the $R$ registers of Appendix~\ref{app:gkv}, it forms the $E$ extra keys of Section~\ref{app:step-cost-sec}. Unlike the registers, which take a fixed share of each window's keys at every scheduled size, a ring with a fixed maximum width takes a growing share as the window shrinks. The two are complementary, since the ring provides a dense band of immediate neighbors while the registers provide a sparse sample of the rest of the image.

\section{Global-KV Bridge}
\label{app:gkv}

Phase~I establishes a coherent layout, but in Phase~II each window attends only to the tokens inside it and the thin ring of the halo. A window therefore cannot see whether a small object already exists elsewhere in the image, and may generate a second copy (Figure~\ref{fig:ablation-halogkv}). The global-KV bridge addresses this by giving every window a compact summary of the image outside it, added as extra keys and values within the same softmax as its local keys.

\subsection{What Is Appended}
\label{app:gkv-append}

Consider a window with $n = w^2$ image tokens. After rotary position embeddings are applied, we append $R$ extra keys and values, taken from image tokens outside the window, and leave the queries unchanged. The attention call then has queries $[\text{text} \mid \text{window}]$ and keys $[\text{text} \mid \text{window} \mid \text{halo} \mid \text{registers}]$. Since the local and global keys share one softmax, each query sets its own balance between them, without a separate global branch or an additional merge weight.

The $R$ registers are actual tokens rather than pooled averages. As in Phase~I, keys share a large common component, so averaging many of them keeps this component and shrinks the rest. The resulting keys point in nearly the same direction and can absorb attention without carrying distinctive information. We therefore take $K$ and $V$ from the same selected token. $K$ is taken after RoPE, so each register keeps the true position of its token in the image, while $V$ is not rotated, as usual. Placing the registers at the origin, or on a separate coarse grid, would not match the positions of the queries, since all keys in a single attention call are compared with the same rotated queries. Because the registers keep their true positions, a query and a register can be as far apart as the full image, farther than any distance seen during training. The bridge therefore reintroduces some unseen distances, but only for the few register keys in each window. Moreover, the global layout is already established by Phase~II, and we do not observe harmful effects from these distances at this stage of denoising. After rotary position embeddings are applied, we append $R$ extra keys and values, taken from image tokens outside the window, and leave the queries unchanged. Tokens inside the window are excluded, since the window already attends to all of them, so registers drawn from inside it would only repeat this information more sparsely while using up part of the register budget.

\vspace{-5pt}
\subsection{Which Tokens Are Selected}
\label{app:gkv-select}
\vspace{-3pt}

Let $k_i$ be the unrotated key of image token $i$, and let $\bar k = \frac{1}{HW}\sum_j k_j$ be the mean key over the image. Each token receives an importance score
\begin{equation}
\label{eq:gkv-score}
s_i
\;=\;
\bigl\|k_i - \bar k\bigr\|_{2}^{2}.
\end{equation}
A large $s_i$ means the token differs from the rest of the image, which is the content a distant window most needs to know about to avoid generating a second copy.

Simply selecting the $R$ tokens with the highest scores across the whole image does not work well. The selection can concentrate on a single striking object, and empty regions never score highly, so they are never represented. To avoid duplicating content, however, a window needs to know both what exists elsewhere in the image and where nothing exists. AReA therefore distributes the $R$ registers over an $8 \times 8$ grid of cells covering the image and, within each cell, selects the highest-scoring tokens outside the querying window. Every window still receives exactly $R$ distinct registers, so all windows keep the same key layout and the batched attention call is unchanged.

\begin{figure}[t]
  \centering
  \includegraphics[width=\textwidth]{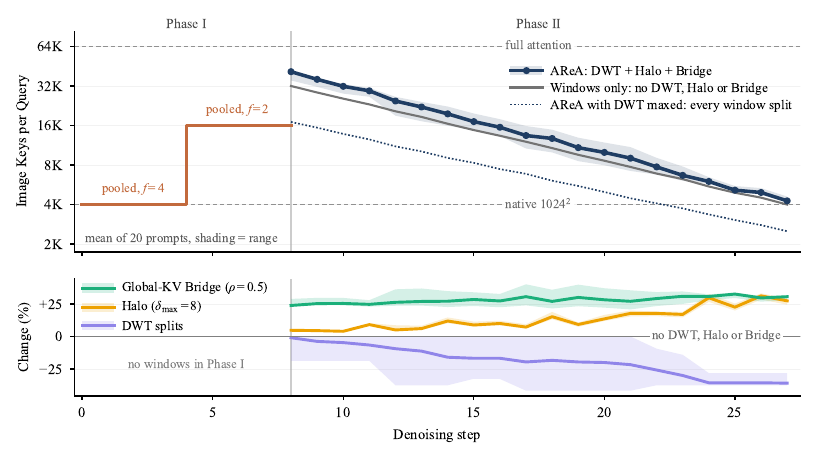}
  \vspace{-20pt}
  \caption{\textbf{Key budget per denoising step.} Number of image keys each query attends to on FLUX.1 at $4096 \times 4096$, averaged over 20 prompts (shading shows the range). \textit{Top}: In Phase~I, the pooled keys match the token count of the native $1024^2$ resolution for $f = 4$ and four times that for $f = 2$. In Phase~II, the windows shrink from about 32K keys to the native 4K, and full AReA stays within about 30\% of plain windowing. No query attends to more than about 40K of the 64K image keys at any step. \textit{Bottom}: Change in keys per query relative to plain windows. The global-KV bridge adds a roughly constant share of about 25\%, the halo takes a growing share as windows shrink, and DWT splits remove up to 35\% late in Phase~II.
  \vspace{-10pt}
  }
  \label{fig:key-budget}
\end{figure}

\vspace{-5pt}
\subsection{How Many Registers}
\label{app:gkv-count}
\vspace{-3pt}

A constant $R$ would give every window the same number of registers throughout Phase~II, so as windows shrink, the registers would make up a growing share of their keys. Global context, however, matters most early in Phase~II, when duplicated objects are most likely to form, and least late, when only texture is refined. We therefore scale the number of registers with the window size, so that larger windows receive more registers, with a fixed ratio of registers to window tokens at every step,
\begin{equation}
\label{eq:gkv-R}
R
\;=\;
(\rho\, w)^{2}.
\end{equation}
We use $\rho = 0.5$ for all models, so each window has one register for every four of its own tokens. Since the window schedule is chosen using $C(w)$ alone, adding the halo and the registers can make the first Phase~II steps more expensive than full attention, since the windows are largest and overlap the most at the start of Phase~II. We therefore check the cost of every Phase~II step after these additions. If it exceeds the cost of full attention, we reduce the halo and register keys until it no longer does. As a result, no Phase~II step costs more than full attention.

Figure~\ref{fig:key-budget} shows how the three additions change the number of keys over a full run. The bridge adds about a quarter to each window's keys throughout Phase~II, as set by Eq.~\ref{eq:gkv-R}. The halo adds little while the windows are large and grows to about a quarter as they approach the floor. The DWT splits offset both, removing up to a third of the keys late in Phase~II, when the gate opens most. As a result, the number of keys per query stays within about 30\% of the plain windowed schedule and below the full image at every step.

\begin{figure}[t]
  \centering
  \includegraphics[width=\textwidth]{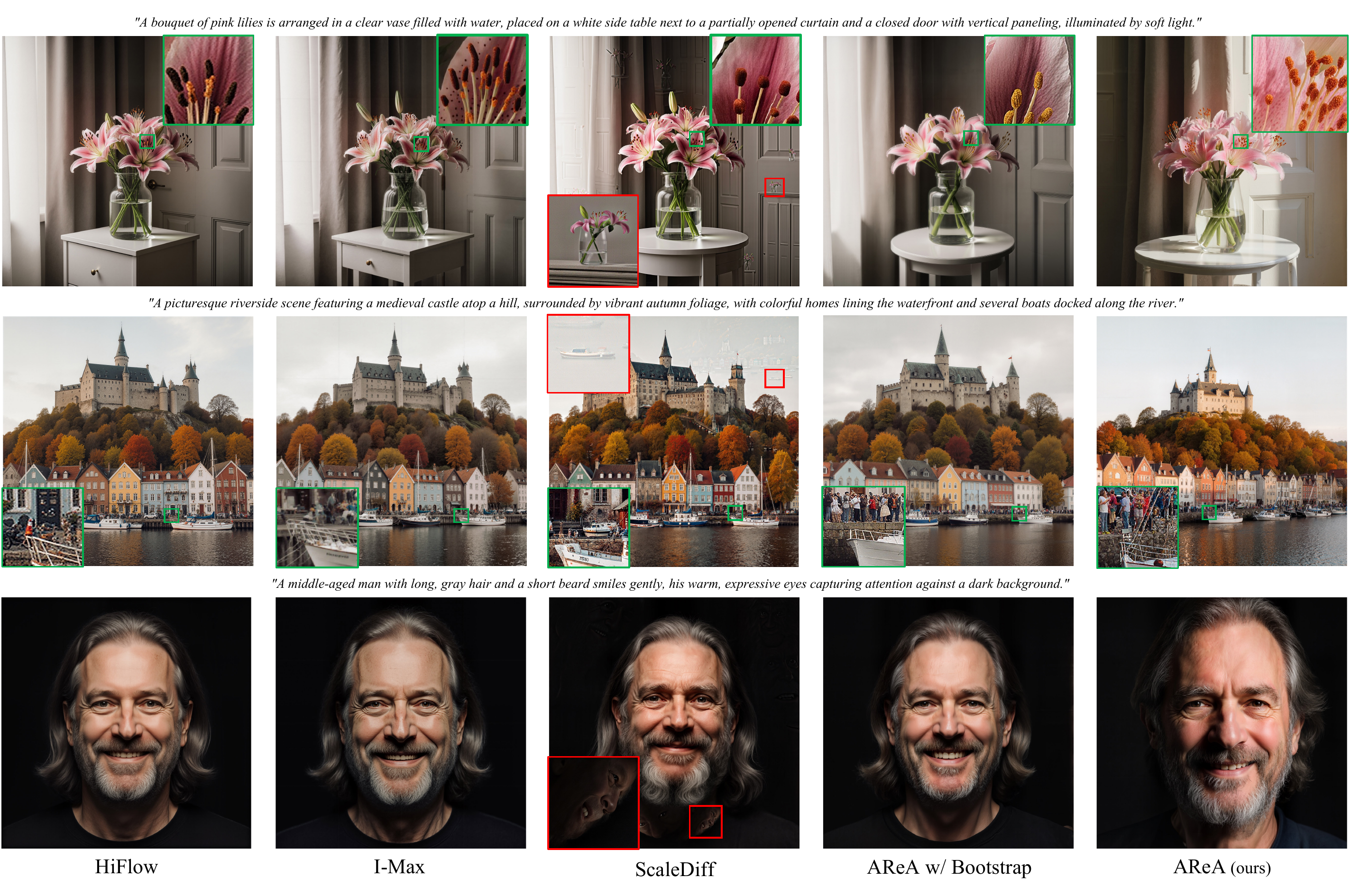}
  \vspace{-20pt}
  \caption{\textbf{Comparison with Multi-Stage Methods on FLUX.1 at $\mathbf{4096 \times 4096}$}. Multi-stage methods and AReA-Bootstrap inherit the composition of the same base-resolution image and differ mainly in local detail. AReA with Phase I (default) sets its own layout at the target resolution and renders finer detail (green insets). ScaleDiff introduces repetitive content in low-detail regions (red insets).
  \vspace{-10pt}
  }
  \label{fig:multistage}
\end{figure}

\section{Comparison with Multi-Stage Methods}
\label{app:multistage}

Multi-stage methods such as HiFlow~\citep{hiflow}, I-Max~\citep{imax}, and ScaleDiff~\citep{scalediff} first generate an image at the model's native resolution and then use it, or its sampling trajectory, to guide generation at the target resolution. Their high-resolution output is therefore anchored to the base image. Its layout is fixed at the native resolution, and the target-resolution pass mainly adds local detail. In this sense, these methods are closer to generative super-resolution than to direct high-resolution synthesis~\citep{resdit, sega}, which AReA and the RoPE-based methods perform end-to-end at the target resolution. We therefore compare against them separately on FLUX.1, the backbone used in their official implementations.

To compare with these methods under the same setting, we also evaluate AReA in a \emph{bootstrap} mode that follows the same paradigm. Phase~I is replaced by a generation at the native resolution. The resulting latent is upsampled to the target latent size without decoding and re-noised to the noise level at which Phase~II begins ($\sigma^\star = 0.89$ on FLUX.1), and the remaining 20 steps are denoised with Phase~II unchanged.

As shown in Table~\ref{tab:multistage}, AReA-Bootstrap is the fastest method and outperforms the multi-stage methods on ImageReward, Aesthetic, CLIP, and FID, while trailing ScaleDiff on the remaining no-reference quality metrics. This suggests that Phase~II is an effective refiner even when the layout comes from a base low-resolution image. Full AReA, however, achieves the best score on every quality, alignment, and fidelity metric. Figure~\ref{fig:multistage} illustrates the difference. All multi-stage outputs, including AReA-Bootstrap, share nearly the same composition because they inherit the layout of the same base image, whereas full AReA forms its own layout at the target resolution and renders finer detail.

\input{tables/multi-stage_table}

\begin{figure}[t]
  \centering
  \includegraphics[width=\textwidth]{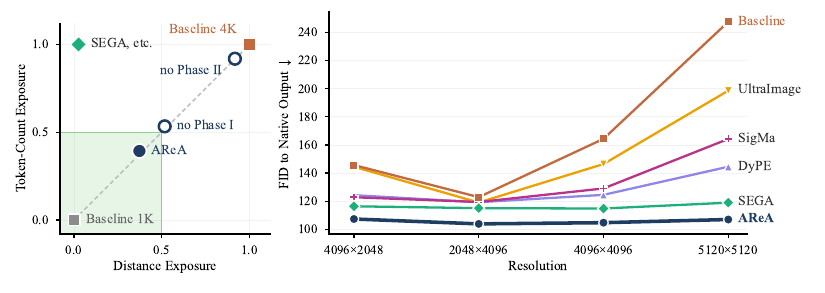}
  \vspace{-20pt}
  \caption{\textbf{Out-of-Distribution Exposure and Its Consequence on FLUX.1.} \emph{Left}: distance and token-count exposure at $4096^{2}$, averaged over denoising and scaled so that $0$ is the model at its native $1024^{2}$ and $1$ is the vanilla model at $4096^{2}$. RoPE-modifying methods (like SEGA) remove only the distance cause, while AReA reduces both. \emph{Right}: FID of each method's outputs to the model's own native-resolution outputs; lower is closer to the model's intended and trained distribution. AReA stays lowest and flat as resolution grows.
  \vspace{-10pt}
  }
  \label{fig:ood}
\end{figure}

\section{Out-of-Distribution Exposure}
\label{app:ood}

A key factor in high-resolution generation is how well a method keeps the model within the conditions it was trained on. This section describes how attention departs from these conditions when a diffusion transformer runs above its training resolution, measures how far each method departs from them, and shows the effect on the output. The causes and AReA's response to them apply to any backbone with rotary position embeddings, and we carry out the measurements on FLUX.1.

\subsection{Two Ways a Large Image Leaves the Training Distribution}
\label{app:ood-causes}

Consider a model trained on a $P \times P$ patch grid and run on an image $k$ times larger per side. Its attention then departs from the training distribution in two distinct ways.

\textbf{Distance.} Query-key offsets grow to about $kP$ patches per axis, while training never saw more than $P$. RoPE turns each offset into a rotation angle in every frequency band, and in the low-frequency bands, whose period is longer than the training range, large offsets produce angles the model has never seen.

\textbf{Token count.} Every query attends to $k^2$ times as many keys as in training. Softmax spreads its weight over all of them, which flattens attention even when every offset is familiar.

Methods that adjust RoPE, such as SEGA~\citep{sega}, DyPE~\citep{dype}, and UltraImage~\citep{ultraimage}, rescale the rotary frequencies so that large offsets map back onto familiar angles. This addresses distance only. Every query still attends to every key in the image, so the token count is the same as in the baseline. AReA addresses both. In Phase~I, the pooled grid starts at the native size, which keeps both the offsets and the number of keys within the training range, and grows beyond it only in the final Phase~I steps. In Phase~II, each window limits the offsets and the number of keys to its own size, which shrinks to the native size by the end of the phase. The bridge registers are the one exception, since they keep their true positions (Appendix~\ref{app:gkv-append}).

\subsection{Measuring Exposure}
\label{app:ood-exposure}

To separate the two causes, we measure each on its own. For every method and every denoising step $t$, we quantify how much of its attention falls outside the training conditions for each cause. Let $\mathcal{A}_t$ be the set of query-key pairs the method attends to at step $t$, and let $\Delta$ be the offset between the query and the key of a pair. The \emph{distance exposure} $D_t$ is the fraction of pairs in $\mathcal{A}_t$ and rotary frequencies whose rotation angle $\hat\omega\,\Delta$ lies outside the range $[-\omega(P-1),\,\omega(P-1)]$ covered on the training grid. Here $\hat\omega$ is the frequency the method uses, which equals $\omega$ unless the method rescales RoPE. The \emph{token-count exposure} $K_t = \log_2(n_t / P^{2})$ compares the number of keys $n_t$ each query attends to with the number seen in training. Both depend only on which pairs are attended, not on the attention weights.

Steps do not contribute equally to the final image, so we average each quantity over the trajectory with weights proportional to the noise the step removes, $\sigma_t - \sigma_{t+1}$. We then express each average relative to two references, the model at its native resolution, where $D = K = 0$ by construction, and the model at the target resolution with no method, which we set to $1$. A value of $0.5$ therefore means that a method has removed half of that cause.

Figure~\ref{fig:ood} (left) places each method on these two axes for FLUX.1 at $4096^2$ ($P = 64$, $k = 4$). Rescaling the frequencies brings almost every rotation angle back into its trained range, but each query still attends to the full image, so the token count is the same as in the baseline. AReA sits at $(0.37, 0.39)$, removing about $60\%$ of each cause, and both phases contribute. In Phase~I, the pooled grid begins at exactly the native size, so the early steps, which decide the layout, run in distribution on both axes. In Phase~II, each window limits both how far a query can look and how many keys it sees, and the windows shrink to the native size as denoising proceeds. The remaining exposure comes mainly from the larger windows early in Phase~II and from the global-KV bridge, whose tokens keep their true positions in the image. Removing either phase moves AReA along the diagonal toward the baseline.

\subsection{Measuring the Consequence}
\label{app:ood-fid}

If exposure causes artifacts, a method's outputs should drift away from what the model produces within its training distribution. We measure this drift as the FID between each method's outputs and the same model's outputs at its native resolution with the same aspect ratio, using the same prompts and seeds. Since the reference is the model's own output rather than real images, this FID measures how far a method pulls the model away from its own distribution, not the general quality of the images. FID also resizes every image to $299^2$, so it reflects global statistics such as layout and color rather than fine detail. Since each resolution has its own reference set, FID values are most meaningful when comparing methods at the same resolution, and trends across resolutions should be read with this in mind.

Figure~\ref{fig:ood} (right) shows the results on FLUX.1. AReA has the lowest FID at every resolution and stays nearly flat across resolutions. The baseline drifts further as the image grows. The RoPE-based methods stay below the baseline but also drift upward at the two largest resolutions. SEGA, the method closest to native on the distance axis, stays above AReA at every resolution.

\subsection{Why Adjusting RoPE Is Not Enough}
\label{app:ood-rope}

On FLUX.1, SEGA essentially removes the distance cause, yet its outputs remain further from the model's own distribution than AReA's at every resolution, even though AReA retains more distance exposure. The other RoPE-based methods also drift further as the image grows. What these methods leave unchanged is the token count, which grows with the square of the side for any backbone ($16\times$ at $4096^2$ and $25\times$ at $5120^2$ for a model trained at $1024^2$). Among the methods we compare, AReA is the only one that reduces both causes, and its outputs stay closest to the model's own distribution at every resolution. This is consistent with the token count being a major source of the remaining out-of-distribution behavior.

\section{HRGen-1K Benchmark}
\label{app:hrgen}

Existing evaluation sets for high-resolution generation are relatively small, such as Aesthetic-4K~\citep{aesthetic4k} with 195 prompts, and cover a limited range of scene complexity. Inspired by SEGA~\citep{sega} and JoLT~\citep{jolt}, we introduce \textbf{HRGen-1K}, a benchmark of 1,000 curated prompts generated with GPT-5.6 Sol and Claude Opus 5. The prompts are designed to test the two properties that high-resolution generation must balance, a coherent global structure across a large canvas and fine detail at the local level, under diverse scenes, styles, and prompt formats. To this end, many prompts describe complex scenes with multiple interacting subjects, dense layouts, and unusual situations. Such prompts expose failures that simple single-subject prompts rarely reveal, such as duplicated subjects, distorted structures, and details that remain blurry or fragmented as the canvas grows.

\paragraph{Content and diversity.}
HRGen-1K spans a broad range of scene categories. Natural scenes cover landscapes, coastlines, and weather phenomena, while built environments range from architectural exteriors to detailed interiors. Human-centered prompts include portraits, fashion, street crowds, and people at work, and object-centered prompts feature food, still life, macro textures, and vehicles. The benchmark further includes wildlife, historical settings, narrative moments, imaginative worlds, surreal concepts, and complex illustrated compositions. Many prompts pair a scene description with an explicit fine-detail constraint and a style directive, such as requiring individually resolved water droplets or species-accurate fur, while others are free-form descriptions. About 35\% of the prompts request photographic styles and 21\% request artistic styles such as illustration, painting, or paper craft. The remaining 43\% leave the style unspecified, evaluating how well a model generates high-resolution images without any stylistic guidance.

\begin{figure}[h]
\centering
\includegraphics[width=\linewidth]{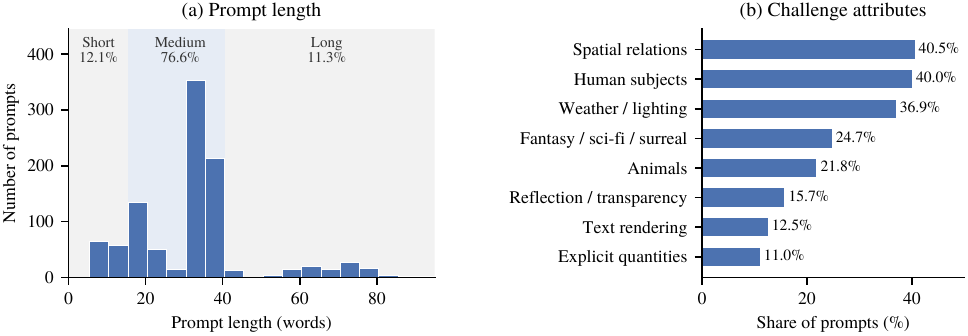}
\caption{\textbf{Statistics of HRGen-1K}. \emph{left}: Distribution of prompt lengths, grouped into short ($\leq 15$ words), medium (16-40 words), and long ($> 40$ words) prompts. \emph{Right}: Share of prompts containing common attributes, estimated via keyword matching. Attributes are not mutually exclusive.}
\label{fig:hrgen-stats}
\end{figure}

\paragraph{Prompt length.}
HRGen-1K also covers a diverse range of prompt lengths, from 9 to 91 words with an average of 32.5 words (Figure~\ref{fig:hrgen-stats}(a)). Short and long prompts pose different challenges. Short prompts, which make up 12.1\% of the benchmark, provide little information about the scene, so the model must infer plausible details on its own to fill a large canvas coherently. Long prompts, which make up 11.3\%, describe many subjects and attributes at once, making it harder to follow every instruction and increasing the risk of omitted or mixed objects. All 1,000 prompts are unique, with a vocabulary of 4,867 distinct words.

\paragraph{Challenge attributes.}
Beyond scene diversity, HRGen-1K targets known failure modes of generative models (Figure~\ref{fig:hrgen-stats}(b)). A large share of prompts involve explicit spatial relations, human subjects whose anatomy must remain coherent, and specific weather or lighting conditions. The benchmark also includes 125 text-rendering prompts, ranging from single words on reflective or transparent surfaces to full sentences of up to 18 words that must preserve punctuation and numbers, as well as prompts with explicit object quantities, reflections, and transparency.

We hope HRGen-1K serves as a more challenging and comprehensive benchmark for evaluating future high-resolution generation methods. We will publicly release the full benchmark.

\section{Extended Related Work}
\label{app:relatedwork-apdx}
\vspace{-5pt}

In this section, we provide background on diffusion models and review prior work on high-resolution image generation in more detail, highlighting how our approach differs from existing methods.


Text-to-image (T2I) generation has advanced rapidly with diffusion models~\citep{ddpm, ddim}, which produce high-quality images that closely follow textual prompts. A key step was latent diffusion~\citep{latentdiff}, which reduced computational cost and enabled models such as Stable Diffusion and SDXL~\citep{sdxl}. More recently, Rectified Flow~\citep{liu2022flow} and Diffusion Transformers (DiTs)~\citep{dit} have further improved quality and scalability, replacing the U-Net backbone~\citep{unet} with transformer architectures that capture global context and scale effectively with model and data size, enabling powerful text-to-image models such as FLUX.1~\citep{flux}, FLUX.2~\citep{flux2}, Qwen-Image~\citep{qwen} and Ideogram 4.0~\citep{ideogram}.


High-resolution image generation aims to synthesize large, coherent images beyond a model's native training resolution. Naively applying pretrained diffusion models at higher resolutions often leads to repetitive patterns and structural distortions. Several works address this by training or fine-tuning on high-resolution data~\citep{guo2024make, hoogeboom2023simple, liu2024linfusion, teng2024relay, aesthetic4k}. For example, URAE~\citep{urae} proposed parameter-efficient adaptation, LWD~\citep{lwd} used wavelet-guided supervision to enhance fine details, UltraImageGen~\citep{ultraimagegen} combined hierarchical local attention with low-resolution global guidance, and JoLT~\citep{jolt} jointly denoised low- and high-resolution latent trajectories to balance global coherence and local detail. 

This has motivated training-free approaches that enable models trained at their native resolution to operate effectively at higher resolutions without additional data or fine-tuning. Instead of updating model weights, these methods~\citep{multidiffusion, fouriscale, upsample-guidance, jin2023training} modify the inference process, either by guiding high-resolution sampling with a base-resolution generation or by adjusting internal components such as attention and positional encoding. Some works~\citep{demofusion, accdiffusion} subdivide the target high-resolution images into overlapping trained-resolution patches, which are processed individually and then stitched together. MultiDiffusion~\citep{multidiffusion} fuses diffusion paths over spatial windows to enable spatially controlled generation. DiffCollage~\citep{diffcollage} similarly composes large content through parallel diffusion generation, while SyncDiffusion~\citep{syncdiffusion} synchronizes overlapping diffusion windows to decrease window incoherence for separate patches. However, the required overlap substantially increases computational cost and can lead to object repetition artifacts.

With the shift toward Diffusion Transformers, training-free high-resolution methods have largely adopted two-stage pipelines, in which a base-resolution generation guides sampling at the target resolution. Some methods~\citep{diffusehigh, upsample-guidance} upsample the base-resolution image and refine it with techniques such as SDEdit~\citep{sdedit}, relying on the model's ability to generate strong local details at higher resolutions. I-Max~\citep{imax} treats the base-resolution flow as a projection of the high-resolution flow and exploits the linear interpolation property of rectified flows to construct dynamic guidance that steers high-resolution sampling. HiFlow~\citep{hiflow} extends this idea by constructing a virtual reference flow from the full base-resolution trajectory, providing initialization, direction, and acceleration guidance that aligns high-resolution sampling with coarse structure while enhancing fine details. HiDiffusion~\citep{hidiffusion} resizes feature maps in the outer blocks of the U-Net so that deeper blocks operate at the training resolution, and replaces global self-attention with shifted-window attention to improve efficiency. ResMaster~\citep{resmaster} uses a low-resolution reference to guide patch-wise high-resolution generation, aligning low-frequency components to preserve global structure while using reference-derived image and enriched text prompts to refine local details. DemoFusion~\citep{demofusion} generates high-resolution images through progressive upscaling, using skip residuals from the base-resolution generation to preserve global structure and dilated sampling to improve global coherence across local patches. DiffuseHigh~\citep{diffusehigh} upsamples the base-resolution image and re-denoises it, using the Discrete Wavelet Transform (DWT) to preserve its structure. ScaleDiff~\citep{scalediff} upsamples the base-resolution image and refines it with SDEdit~\citep{sdedit} to add local details at the target resolution. It combines Neighborhood Patch Attention with a frequency-mixing strategy that merges low-frequency content from latent-space upsampling with high-frequency content from RGB-space upsampling, and applies latent-space structure guidance to preserve global structure while recovering fine details.

While effective, these methods depend on base-resolution guidance and introduce additional complexity into the denoising process. Since the high-resolution output is anchored to the base image, its quality is bounded by that image, and any errors in the initial generation propagate to the final result. High-resolution generation is not simply a matter of producing more pixels, as the level of detail, structure, and composition must also scale with the resolution and remain coordinated across the image. As a result, many methods reach the target pixel count but fail to deliver a corresponding gain in detail fidelity, producing images that are larger but not truly richer in detail. Thus, matching the target pixel count alone does not necessarily yield high-quality images with faithful fine details.

To overcome the limitations of multi-stage approaches, more recent direct-inference methods generate high-resolution images end-to-end, aiming to unlock the capabilities already present in pretrained models by modifying their internal behavior at inference time. This is typically achieved by adjusting the denoising process or key components such as positional encodings and attention, allowing the model to operate beyond its training resolution without additional training. For example, ScaleCrafter~\citep{scalecrafter} attributes repetition and structural artifacts at higher resolutions to the limited receptive field of convolutional kernels and addresses this by dilating convolutions in the U-Net at inference time, though this may distort structure. ResDiT~\citep{resdit} applies global attention with scaled positional embeddings in parallel with patch-wise attention, preserving the overall layout while recovering local details, and merges their outputs through Gaussian-weighted blending and spectral fusion to combine low-frequency structure with high-frequency detail. 

 Another line of work approaches the problem through the lens of RoPE length extrapolation. Modern DiTs encode spatial positions with Rotary Position Embeddings (RoPE)~\citep{rope}, and generating at higher resolutions introduces positions beyond the training range, producing rotation angles the model has never encountered. This closely mirrors context extension in large language models (LLMs), where RoPE frequencies are rescaled to handle sequences longer than those seen during training~\citep{yarn, longrope}. Building on these ideas, image-domain methods typically recalibrate RoPE's rotary frequencies along each spatial axis, so that positions at the target resolution remain within the distribution the model was trained on. DyPE~\citep{dype} introduces a step-dependent positional extrapolation schedule that follows the progression of denoising process, preserving low-frequency global structure early and gradually relaxing positional constraints to recover high-frequency details. UltraImage~\citep{ultraimage} attributes repetition to mid-band RoPE components whose periods align with the training resolution and recursively adjusts them to span the target resolution; it further uses attention entropy to sharpen local patterns while preserving broad global attention. SigMa~\citep{sigma} extends dynamic positional modulation with an axis-wise, scale-adaptive sigmoid schedule that adjusts both the timing and rate of the transition from structural formation to detail refinement. SEGA~\citep{sega} analyzes the spatial-frequency energy of intermediate latents at each denoising step and maps it to per-dimension RoPE scaling, strengthening underrepresented frequency bands while relaxing already dominant ones and using spectral flatness to control the overall correction.
 
Despite their progress, these methods are tied to RoPE-based positional adjustments and inherit the constraints of length extrapolation. Since they modify architecture-specific components, they do not readily transfer to models with different positional encodings, and frequency recalibration alone cannot fully prevent out-of-distribution behavior at extreme scales, often resulting in artifacts and degraded detail fidelity at high resolutions. Moreover, these methods still apply full global attention over all tokens, leaving the quadratic cost of high-resolution generation unaddressed. As a result, balancing local detail, global coherence, and computational cost remains challenging, especially as resolution grows. In contrast, our method generates directly at the target resolution without base-resolution guidance or RoPE extrapolation. AReA adapts the attention receptive field across denoising steps, moving from global to local attention as generation progresses from coarse structure to fine detail. This keeps the model close to its training regime, allowing it to remain stable at resolutions far beyond the training range while reducing attention cost, without retraining or architecture-specific components.

\begin{figure}[t]
  \centering
  \includegraphics[width=0.9\textwidth]{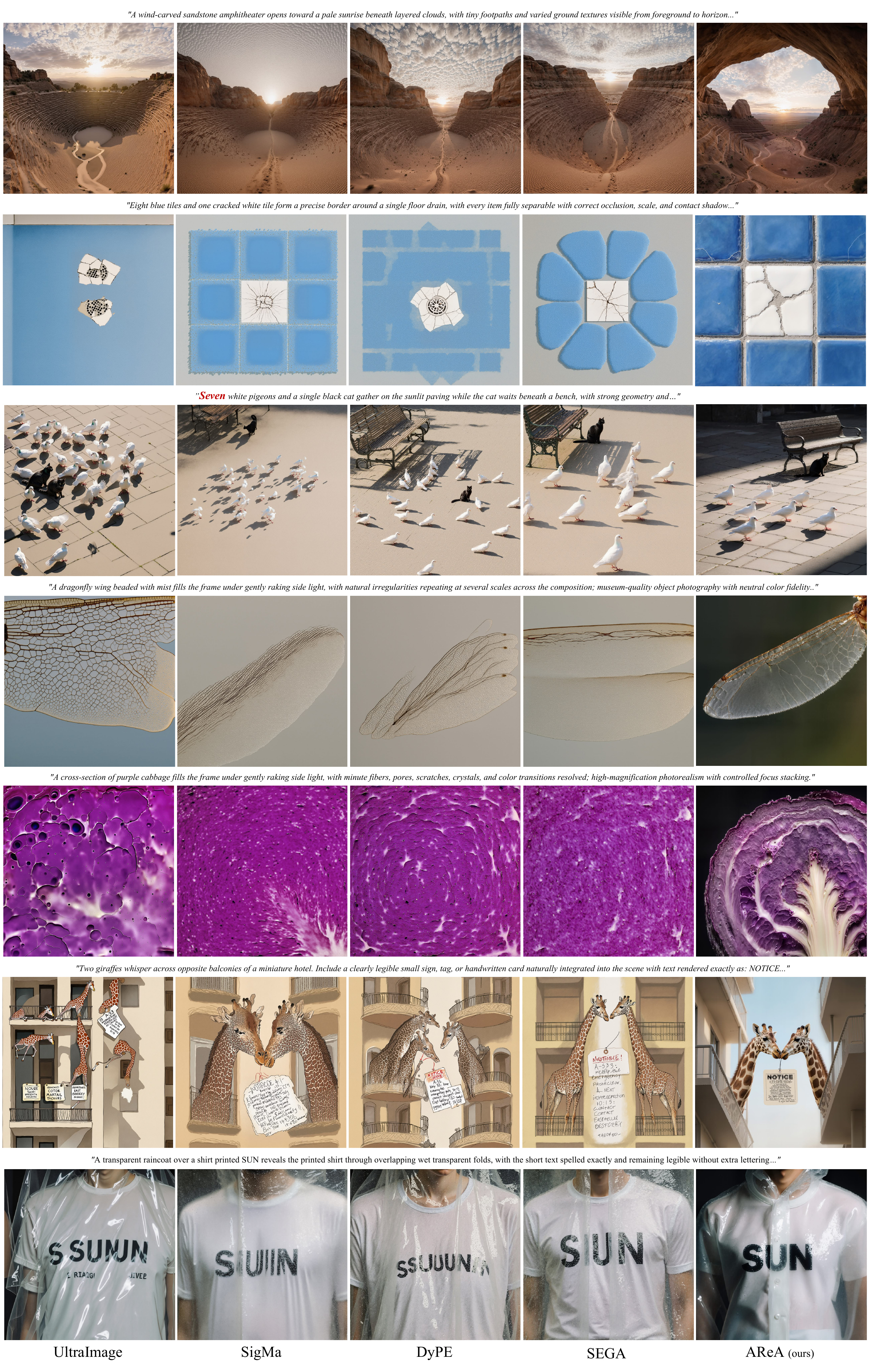}
  \vspace{-5pt}
  \caption{\textbf{Additional Comparison on FLUX.1 at $\mathbf{4096 \times 4096}$ on HRGen-1K}. Existing methods warp scenes into symmetric, fisheye-like layouts, reduce object-centric prompts to zoomed-in texture, miscount objects, and garble rendered text. AReA preserves scene composition, object counts (highlighted in red), and legible text.
  \vspace{-10pt}
  }
  \label{fig:comparison-hrgen1k}
\end{figure}

\begin{figure}[t]
  \centering
  \includegraphics[width=\textwidth]{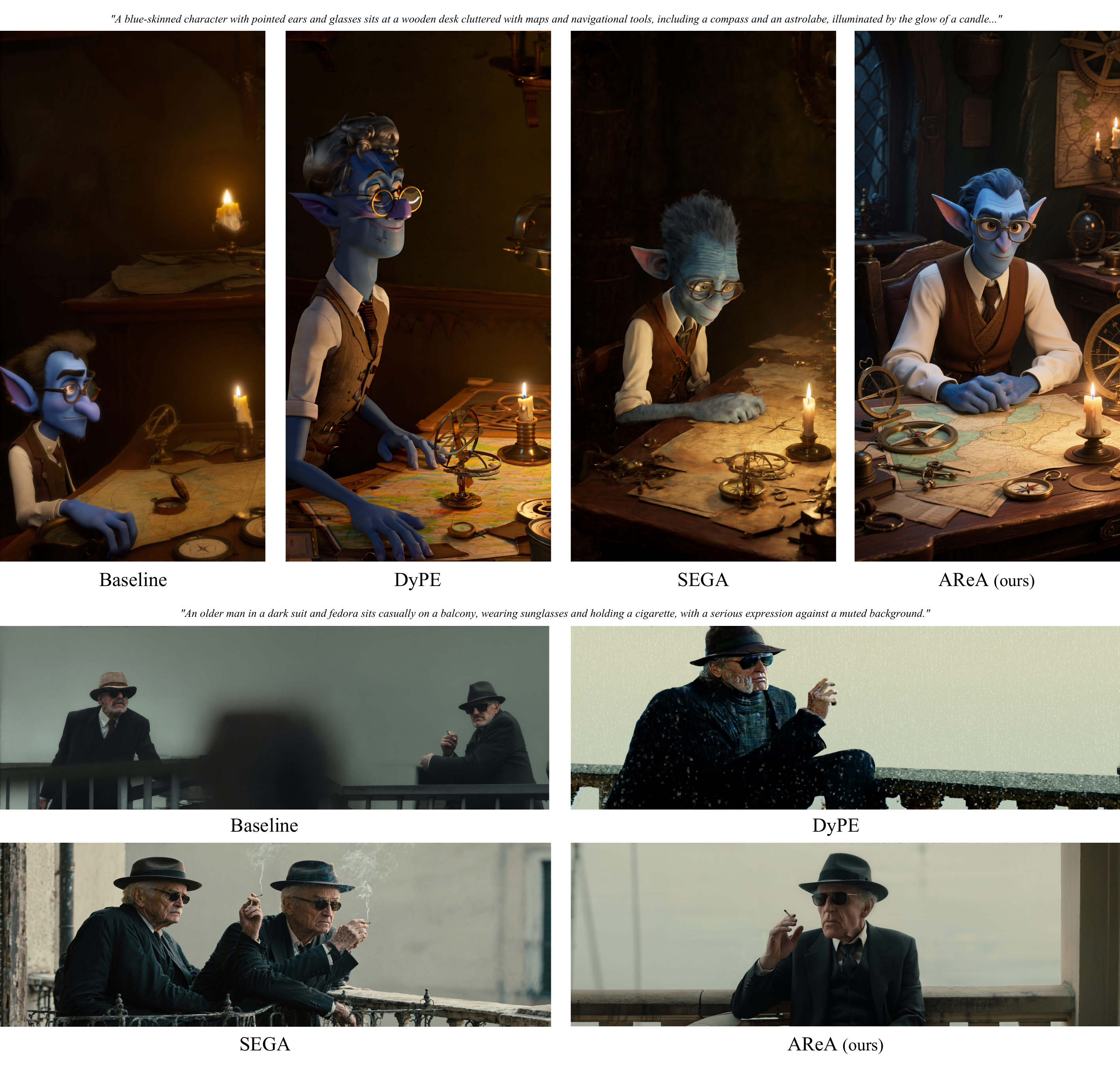}
  \vspace{-20pt}
  \caption{\textbf{Comparison on Qwen-Image at Non-Square Resolutions}. \emph{Top}: $2048 \times 4096$. \emph{Bottom}: $6144 \times 2048$. Full attention misplaces the subject on the tall canvas and duplicates it on the wide one. DyPE stretches the subject and introduces artifacts, and SEGA duplicates the subject in the wide setting. AReA keeps a single, well-proportioned subject in both aspect ratios.
  \vspace{-10pt}
  }
  \label{fig:comparison-ratio}
\end{figure}

\begin{figure}[t]
  \centering
  \includegraphics[width=\textwidth]{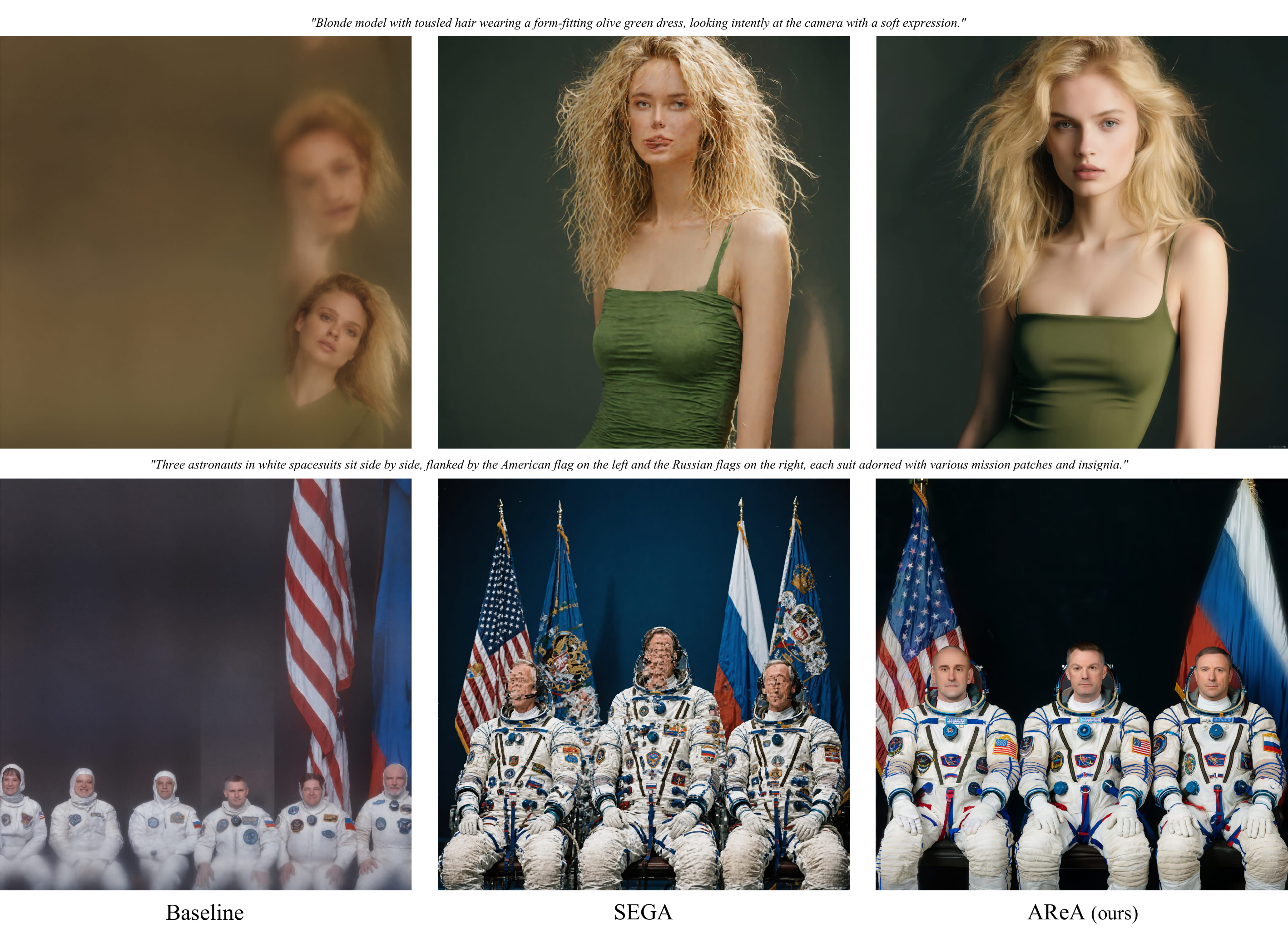}
  \vspace{-20pt}
  \caption{\textbf{Comparison on Qwen-Image at $\mathbf{5120 \times 5120}$}. Full attention blurs the scene and repeats subjects. SEGA keeps the layout but corrupts the details with high-frequency artifacts. AReA produces clean faces and the prompted number of subjects.
  \vspace{-10pt}
  }
  \label{fig:comparison-5k}
\end{figure}

\begin{figure}[t]
  \centering
  \includegraphics[width=\textwidth]{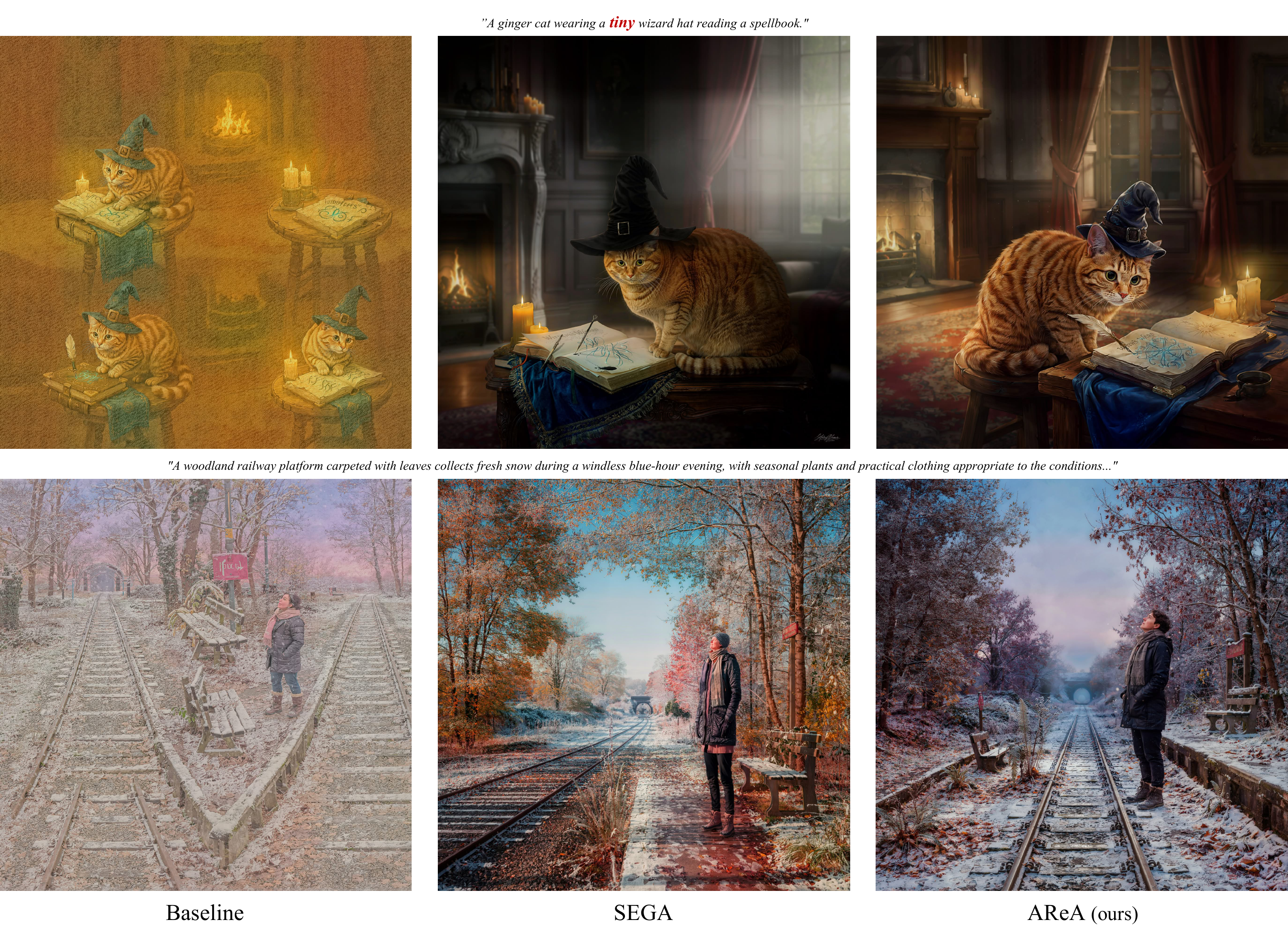}
  \vspace{-20pt}
  \caption{\textbf{Comparison on Ideogram 4 at $\mathbf{6144 \times 6144}$}. Full attention completely collapses, while SEGA fails to keep the layout and add fine detail. AReA excels in both structure and fine detail, showing that it transfers to a different backbone without modification.
  \vspace{-10pt}
  }
  \label{fig:comparison-6k}
\end{figure}

\section{Additional Qualitative Comparisons}
\label{app:qualitative}

We provide further qualitative comparisons against direct-inference methods across backbones, resolutions, and aspect ratios; multi-stage methods are compared separately in Appendix~\ref{app:multistage}. On FLUX.1 at $4096\times4096$, Figure~\ref{fig:comparison-hrgen1k} shows prompts from HRGen-1K that stress composition, object counting, and text rendering, where existing methods distort scene geometry or lose the prompted content while AReA preserves both. On Qwen-Image, Figure~\ref{fig:comparison-ratio} covers portrait ($2048\times4096$) and landscape ($6144\times2048$) canvases, where the failures of other methods grow with the aspect ratio, and Figure~\ref{fig:comparison-5k} shows results at $5120\times5120$. Finally, Figure~\ref{fig:comparison-6k} applies AReA to Ideogram~4 at $6144\times6144$ without any change to the method, illustrating that it transfers across backbones.

%% file: tables/sigma-ablation.tex
\definecolor{defaultcol}{gray}{0.90}
\begin{table}[t]
\centering
\scriptsize
\caption{Ablation of the phase switch step $s$ (Phase~I $\rightarrow$ Phase~II) on FLUX.1 at 4096$\times$4096 resolution on 20 prompts from Aesthetic-4K. The noise level $\sigma$ at the switch is given in parentheses.}
\label{tab:sigma-ablation}
\vspace{0.4em}
\setlength{\tabcolsep}{3pt}
\renewcommand{\arraystretch}{1.05}
\resizebox{\textwidth}{!}{%
\begin{tabular}{@{} l | c | cc>{\columncolor{defaultcol}}ccc | ccccc @{}}
\toprule
 & Too early & \multicolumn{5}{c|}{Stable} & \multicolumn{5}{c}{Too late} \\
\cmidrule(lr){2-2}\cmidrule(lr){3-7}\cmidrule(l){8-12}
Switch step $s$ ($\sigma$) & 2 (0.98) & 4 (0.95) & 6 (0.92) & \textbf{8 (0.89)} & 10 (0.85) & 12 (0.81) & 14 (0.76) & 16 (0.70) & 18 (0.64) & 20 (0.56) & 22 (0.46) \\
\midrule
\multicolumn{12}{@{}l}{\textit{Quality}} \\
ImageReward$\uparrow$ & 0.77 & 1.22 & 1.18 & \textbf{1.25} & \underline{1.24} & \textbf{1.25} & 1.20 & 1.22 & 1.21 & 1.19 & 1.18 \\
Aesthetic$\uparrow$ & 6.03 & 6.20 & \underline{6.22} & \textbf{6.23} & 6.15 & 6.16 & 6.21 & 6.20 & 6.18 & 6.16 & 6.08 \\
CLIP-IQA$\uparrow$ & 0.685 & \textbf{0.712} & \underline{0.705} & \underline{0.705} & 0.696 & 0.686 & 0.671 & 0.658 & 0.628 & 0.579 & 0.528 \\
\midrule
\multicolumn{12}{@{}l}{\textit{Alignment}} \\
CLIP$\uparrow$ & 29.00 & \textbf{30.44} & 30.08 & \underline{30.34} & 30.21 & 30.16 & 30.18 & 30.25 & 30.10 & 30.05 & 29.81 \\
PickScore$\uparrow$ & 22.30 & 22.93 & 23.11 & \textbf{23.24} & 23.16 & \underline{23.20} & 23.01 & 22.98 & 22.92 & 22.76 & 22.54 \\
\bottomrule
\end{tabular}
}
\end{table}

%% file: tables/multi-stage_table.tex
\begin{table*}[t]
\centering
\scriptsize

\caption{Comparison with multi-stage methods on FLUX.1 at 4096×4096 on Aesthetic-4K. \textbf{Bold}: best; \underline{underline}: second-best.}
\label{tab:multistage}

\vspace{0.4em}

\setlength{\tabcolsep}{3pt}
\renewcommand{\arraystretch}{1.05}

\resizebox{\textwidth}{!}{%
\begin{tabular}{p{1.8cm}|cccccccccc}

\toprule
\textbf{Method}
& {Latency (s)$\downarrow$}
& {ImageReward$\uparrow$}
& {Aesthetic$\uparrow$}
& {MUSIQ$\uparrow$}
& {CLIP-IQA$\uparrow$}
& {ARNIQA$\uparrow$}
& {CLIP$\uparrow$}
& {HPS$\uparrow$}
& {FID$\downarrow$}
& {FID$_p\downarrow$} \\

\midrule

HiFlow
& 182
& 1.1710
& 6.3147
& 38.674
& 0.6561
& 0.4811
& 29.3
& 0.2951
& 153.0
& \underline{59.0} \\

I-Max
& 440
& 1.1716
& 6.3043
& 41.086
& 0.6838
& 0.5013
& 29.2
& \underline{0.2956}
& 154.8
& 60.1 \\

ScaleDiff
& \underline{135}
& 1.2339
& 6.2785
& \underline{44.496}
& \underline{0.7066}
& \underline{0.5663}
& 29.2
& 0.2935
& 152.5
& 73.7 \\

\midrule

\textbf{AReA-Bootstrap}
& \textbf{126}
& \underline{1.2658}
& \underline{6.3402}
& 43.224
& 0.7027
& 0.5479
& \underline{29.4}
& 0.2955
& \underline{152.0}
& 61.2 \\

\rowcolor{gray!15}
\textbf{AReA}
& 151
& \textbf{1.3250}
& \textbf{6.3717}
& \textbf{45.217}
& \textbf{0.7132}
& \textbf{0.5942}
& \textbf{29.7}
& \textbf{0.2969}
& \textbf{148.5}
& \textbf{58.2} \\

\bottomrule

\end{tabular}%
}

\end{table*}